\documentclass{article}

\usepackage[main, final]{neurips_2026}

\usepackage[utf8]{inputenc} 
\usepackage[T1]{fontenc}    
\usepackage[plainpages=false,pdfpagelabels]{hyperref}       
\usepackage{url}            
\usepackage{booktabs}       
\usepackage{amsfonts}       
\usepackage{nicefrac}       
\usepackage{microtype}      
\usepackage{xcolor}         
\usepackage{graphicx}       
\usepackage{amsmath} 
\usepackage{multicol}
\usepackage{multirow}
\usepackage{cleveref}
\usepackage{natbib}
\usepackage{placeins}
\usepackage{float}
\setcitestyle{round}
\definecolor{citecolor}{rgb}{0, 0, 0.5}
\hypersetup{breaklinks=true,colorlinks,citecolor=citecolor,linkcolor=citecolor}

\title{Pretraining Data Statistics Shape the Phases of Learning Entity Comparison in Language Models}

\author{%
  \makebox[0.45\textwidth][c]{Yik Siu Chan$^{1,2}$} \\
  \texttt{yiksiu@umd.edu} \\
  \And
  \makebox[0.45\textwidth][c]{Jing Huang$^{3,\dagger}$} \\
  \texttt{hij@stanford.edu} \\
  \AND
  \makebox[0.45\textwidth][c]{Yanai Elazar$^{4,\dagger}$} \\
  \texttt{yanai.elazar@biu.ac.il} \\
  \And
  \makebox[0.45\textwidth][c]{Atticus Geiger$^{2,5,\dagger}$} \\
  \texttt{atticus@goodfire.ai} \\
  \AND
    $^{1}$University of Maryland
    \quad $^{2}$Pr(Ai)$^2$R Group
    \quad $^{3}$Stanford University
   \\
  \AND
    $^{4}$Bar-Ilan University
    \quad $^{5}$Goodfire
}

\begin{document}
\maketitle

\begingroup
\renewcommand{\thefootnote}{\ensuremath{\dagger}}
\footnotetext[0]{Co-senior authors.}
\endgroup

\begin{abstract}
How does data shape language model (LM) behavior throughout pretraining?
We investigate this question through a case study on entity comparison, e.g., \textit{Between France and Brazil, which country is larger?}, across nine comparison tasks spanning five entity types.
We begin with controlled experiments in which we train small LMs (163M parameters) on mixtures of natural text from pretraining corpora and synthetic task data.
Our analysis identifies three distinct phases of learning: (1) an early phase where the LM selects entities by frequency, (2) a middle phase where the LM selects entities by position in the prompt (first vs. last mentioned entity), and (3) a late phase where the LM selects the entity that is the correct answer.
We show that the emergence of these three phases is causally determined by statistical properties of the training data. 
With small amounts of task-specific synthetic data, we observe only the first two phases and the model fails to learn the task; with large amounts, the model jumps directly from the frequency-based heuristic to solving the task correctly.
Moreover, changing the entity frequency distribution from a naturally occurring Zipfian distribution (a small number of entities are very common and the vast majority are rare) to a uniform one weakens the frequency phase and speeds up learning of the task. 
Lastly, we show that the three phases of training dynamics from our controlled experiments appear in the open source model OLMo-2 32B.
Together, our findings demonstrate that properties of pretraining data are causal drivers of heuristic learning and show that small-scale synthetic experiments can predict training dynamics at larger scales.\footnote{Code and data are available at \url{https://github.com/yiksiu-chan/entity-comparison}.}
\end{abstract}
\section{Introduction}
What does a language model (LM) learn throughout pretraining and how do statistical properties of data drive these learning dynamics? 
The relationship between data and model has proven fertile ground for interpretability research, with recent works uncovering how properties of data drive the formation of structured internal representations \citep{merullolinear, karkada2026symmetrylanguagestatisticsshapes, prieto2026correlations}, algorithmic motifs \citep{olsson2022context, chen2024sudden, wutransformers}, generalization capabilities \citep{chan2022data, chen2024sudden,ruis2025procedural, wang2025generalization}, and imperfect heuristics \citep{elazar2022measuring, he2025supposedly, kalai2025language}.
We conduct a case study on learning dynamics in which we study how LMs learn to make factual comparisons between entities, e.g., \textit{Between France and Brazil, which country is larger?}

\begin{figure}[t]
    \centering
    \includegraphics[width=0.98\linewidth]{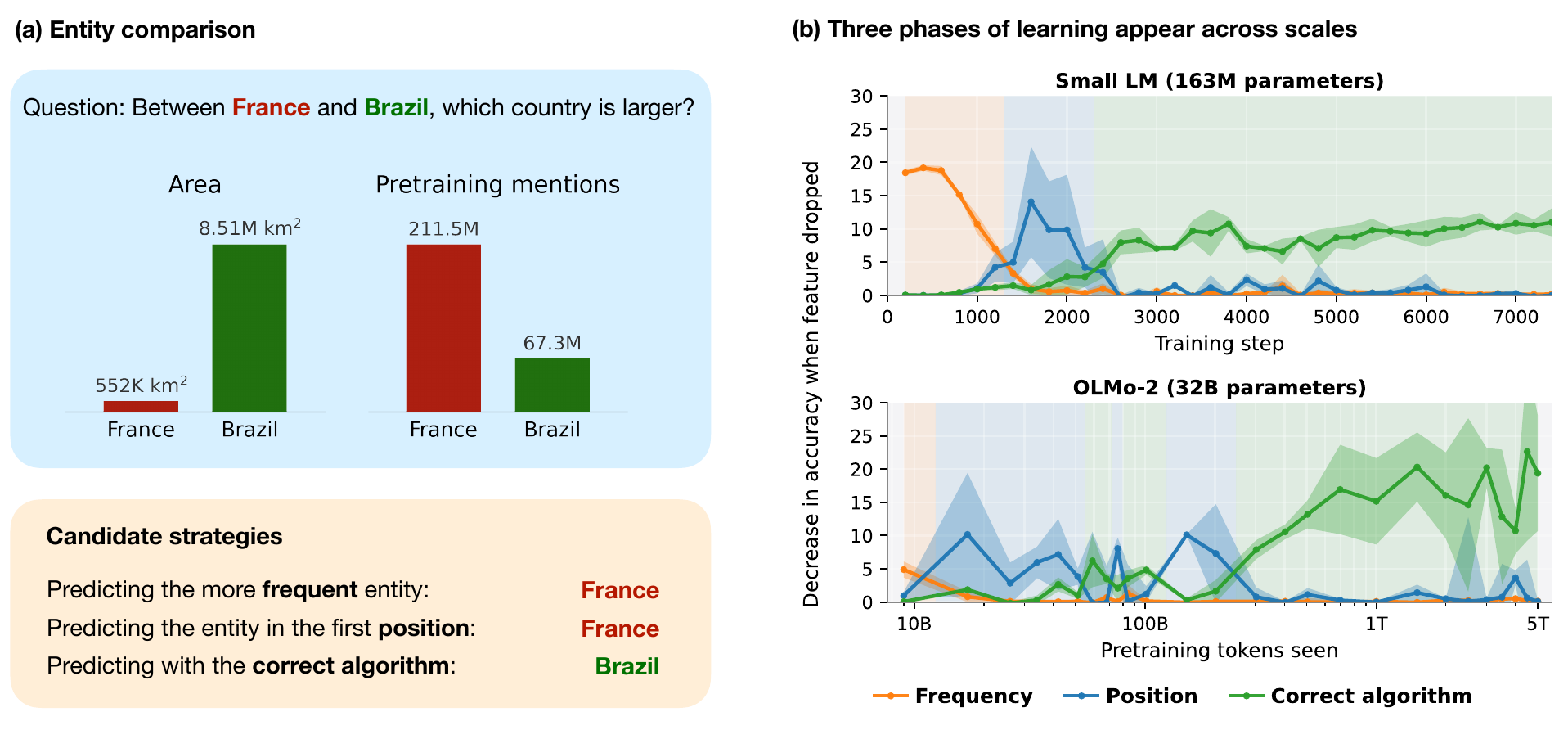}
    \caption{(a) An example of the entity comparison task we use to disentangle the three candidate strategies. The task requires comparing the areas of two countries; the correct answer is Brazil in the given example. However, if the model relies on positional or frequency bias, it would predict France. (b) The importance of each strategy across training checkpoints, measured by the decrease in accuracy when the feature for that strategy is removed. \textbf{The frequency heuristic (\textcolor[HTML]{ff7f0e}{orange}) is most predictive of LM behavior in the earliest checkpoints, but is quickly taken over by the positional heuristic (\textcolor[HTML]{1f77b4}{blue}), and the correct algorithm (\textcolor[HTML]{2ca02c}{green}) appears in late training.} The three-phase structure appears in both the small controlled setting and the large-scale natural text pretraining setting of OLMo-2 models, up to 32B parameters.}
    \label{fig:overview}
\end{figure}

Answering such questions properly requires retrieving the relevant scalar attributes for each entity (e.g., the area of a country), and performing a numerical comparison between those scalar values \citep{elazar-etal-2019-large}.
However, models can instead rely on simpler heuristics, such as the answer's position in the prompt~\citep{zheng2023judging}, or how often the answer is mentioned in the training data~\citep{gottesman2025lment,yang-etal-2025-large}.
Thus, this task provides a natural testbed for understanding (1) \emph{when} different strategies emerge and disappear throughout training and (2) \emph{how} entity frequency in pretraining data shapes such learning dynamics. 

To answer the first question, we characterize model behavior throughout training, using three pieces of information: the difference in scalar attributes for each entity pair (the correct algorithm), entity frequency, and positional bias.
We introduce a controlled pretraining setup in which we train GPT-2-scale models (163M parameters) from scratch.
At each training checkpoint, we use a regression model to identify which features best predict model outputs, allowing us to track the evolution of strategies over time.
Our analysis reveals three distinct phases of learning. First, the LM develops a frequency-based heuristic, outputting the entity that has been seen more often in training thus far. Next, a position-based heuristic takes over where the LM outputs either the first- or last-mentioned entity. Finally, the LM learns the correct comparison algorithm 
(see \Cref{fig:overview}).

We then investigate how properties of the pretraining data drive these learning dynamics (\Cref{sec:training}).
To make the small LM learn the comparison behavior, we supplement natural text from the pretraining corpus with synthetic data, which contains examples of correctly solving the task.
With insufficient synthetic task data, the LM only goes through the first two phases, developing a frequency heuristic then a position heuristic with no further learning. 
Increasing the proportion of task data first strengthens positional bias and would more quickly drive the LM to learn the comparison algorithm. 
In a second experiment, we vary the distribution of entity frequencies. 
Naturally occurring text follows a Zipfian distribution \citep{condon1928statistics}, where a small number of entities are very high frequency while the vast majority of entities are very low frequency; e.g., in the OLMo-2 pretraining corpus, \textit{China} appears 441M times whereas \textit{Luxembourg} appears only 6.8M times. We show that we can shorten the duration of the first two phases and temper the strength of the frequency heuristic by using uniform distributions for the synthetic or natural pretraining data.

Lastly, we then ask whether the same phase structure arises in larger models pretrained on natural text alone, by analyzing checkpoints from the OLMo-2 family (1B, 7B, 13B, and 32B parameters) across all nine tasks (\Cref{sec:olmo}).  
We find the same phases arising in the same order, although which phases appear at all varies with task and model scale. 
This suggests that the observed dynamics are not an artifact of the synthetic setup; the resulting phase structure also characterizes pretraining at scale.
Together, our results identify properties of pretraining data as causal drivers of the strategies LMs adopt, showing that they can control which strategies emerge during training and when.
\section{Setup}\label{sec:methods}
\paragraph{Task formulation.}
The entity comparison task is defined by triples $(d, e_1, e_2)$, where $e_1$ and $e_2$ are entities (e.g., countries) and $d$ is a domain corresponding to a numerical property (e.g., area).
Each entity-domain pair $(d, e)$ is associated with a scalar value $v_{d,e}$.
The algorithm for solving a comparison query $\mathcal{A}(d, e_1, e_2) = \max(v_{d,e_1},v_{d,e_2})$ returns the entity with the higher value.
We study nine comparison tasks spanning five entity types: countries, U.S.\ states, U.S.\ universities, movies, and notable people.
Each comparison task has two directions, where the correct answer is either the larger or the smaller entity.
We perform pretraining on GPT-2-scale models using three tasks (country area, country GDP, and people birth year), and evaluate OLMo-2 models on all tasks.
The full list of tasks is provided in Appendix~\ref{app:tasks}.

\paragraph{Training data.}
We use subsets of the fully open-source OLMo-2 pretraining corpus, \texttt{OLMo-mix-1124} \citep{OLMo20242O2}, which is used primarily for stage-1 pretraining of OLMo-2 models.
Entity mention frequencies in this corpus follow a heavy-tailed Zipfian distribution, which \Cref{fig:freq-distributions}(a) demonstrates for the country entity. 
At the 163M scale, models cannot reliably learn long-tail factual knowledge from natural text alone~\citep{allen2023physics, kandpal2023large}, which we also observed in preliminary experiments.
We therefore add synthetic task supervision data as scaffolding: comparison sentences that exactly match the evaluation template, such as \emph{``By total area, among the two countries A and B, the larger country is''}.
This synthetic data serves as a proxy for the task-relevant signal that larger LMs absorb implicitly from massive pretraining corpora.
We use the full set of entities in both training and evaluation.
This mirrors natural pretraining, where models are exposed to all entities they are later evaluated on, but it also means the LM could memorize the comparisons.
To test this, we run an additional experiment with 20\% of entity pairs held out from the synthetic data in Appendix~\ref{app:holdout}.

\paragraph{Model.} We train transformer language models with the GPT-2 architecture (12 layers, 12 heads, 768-dim embeddings) \citep{radford2019language} from scratch on each data mixture.
Training data is tokenized with the \texttt{allenai/dolma2-tokenizer}, matching the tokenizer used by the OLMo-2 models. 
This ensures token-level frequency counts are directly comparable across models.
We train each model for two epochs with three random seeds.

\paragraph{Evaluation.} 
We extract the teacher-forced logits of the first generated token at which the entities differ to determine the model's choice.
We then denote it as a binary output distribution $P_{\mathrm{LM}}$ over the two entities.
We measure per-pair accuracy at every saved checkpoint.

\subsection{Counting Entity Occurrences in Pretraining Data}\label{sec:counting}
To compute frequency-based features, we track cumulative occurrence and co-occurrence counts of the entity terms up to every saved model checkpoint. 
This accurately reflects how much exposure the model has had to each entity at each point during training.
For entities with multiple name variants (e.g.\ \textit{U.S.A.}, \textit{United States}, \textit{the U.S.}), we aggregate counts across all aliases.
We adapt the \texttt{BatchSearch} method from \citet{merullolinear}.
We define $\textsf{count}_t(e)$ as the number of times entity $e$ appears in the pretraining corpus up to training step $t$, and similarly $\textsf{count}_t(e_1, e_2)$ for co-occurrences. 
Our hypothesis is that entity frequency in the pretraining corpus influences which heuristics a model adopts and the order in which they appear during training. 
Specifically, given a query $(d, e_1, e_2)$, if $\textsf{count}_t(e_1) > \textsf{count}_t(e_2)$ at step $t$, the model can be more likely to predict $e_1$ regardless of whether $v_{d,e_1} > v_{d,e_2}$. 

\begin{figure}[t]
  \centering
  \includegraphics[width=\linewidth]{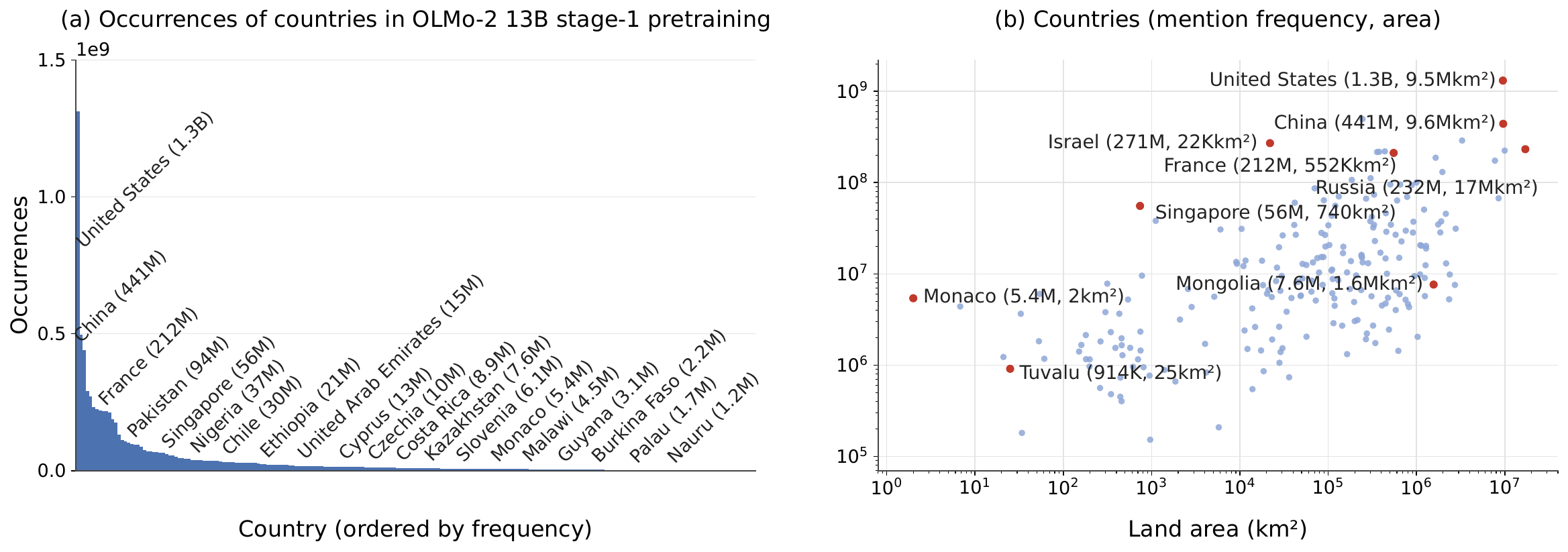}
  \caption{(a) Occurrences of 215 countries (and territories) in the OLMo-2-13B pretraining corpus, which follow a Zipfian distribution with a long tail.  (b) Country frequency correlates with the area attribute (log-log Pearson’s $r=0.62$), meaning that predicting the more frequently mentioned country as the larger one is a viable heuristic the model can adopt.}
  \label{fig:freq-distributions}
  \vspace{-2mm}
\end{figure}

\textbf{Natural pretraining corpora exhibit heavy-tailed Zipfian entity frequency distributions.} In \Cref{fig:freq-distributions}~(a), a small set of countries appears very frequently (e.g., \textit{United States} and \textit{France}), while the vast majority appear far less often. 
This long-tail pattern is typical for words in natural English text \citep{condon1928statistics, zipf2013psycho}. 
In our experiments, we manipulate entity frequency separately in synthetic and natural data to study how this naturally occurring distribution affects learning dynamics.

\subsection{Three Phases of Learning}\label{sec:regression}
To identify which strategies drive LM predictions at each checkpoint during training, we fit a logistic regression model to predict LM outputs from a set of handpicked features, each corresponding to a heuristic or to the correct algorithm. 
By comparing regression accuracy across different feature subsets, we can track which strategies are predictive at each checkpoint. 
In preliminary experiments, we evaluated a wide range of features (full list in Appendix~\ref{app:feature-list}). We found that the best fit uses only three: the frequency, position, and true value of entities. We describe each below, in the order we expect them to dominate over the course of training.

\begin{enumerate}
    \item \textbf{Frequency feature:} We use the log ratio of cumulative occurrences of $e_1$ and $e_2$ in the training corpus up to step $t$: $\log \left(\textsf{count}_t(e_1)/ \textsf{count}_t(e_2)\right)$. A model relying on this feature is using a popularity heuristic, predicting whichever entity it has seen more often. More broadly, we expect this feature to dominate model predictions early in training, before the model has had sufficient exposure to relevant entities.
    
    \item \textbf{Positional bias feature:} Since the two entities are mentioned sequentially, each can appear in either the first or last position. A model in mid-training may be biased to always output the entity in one of these positions, regardless of where the correct answer is. Since the task is a binary choice, we formalize this bias using the regression intercept, which absorbs any systematic preference for the first- or last-mentioned entity.
    Crucially, every entity pair is evaluated in \emph{both} orderings, so any systematic entity-specific bias, such as frequency, scalar value, or tokenization, is counterbalanced across positions. This ensures that the intercept isolates the model's residual preference for a prompt position (see Appendix~\ref{app:position_bias} for verification).

    \item \textbf{Value feature:} We use the log ratio of the true values of $e_1$ and $e_2$ in the question domain $d$: $\log \left(v_{d, e_1}/v_{d, e_2}\right)$. This is the only one of the three features predictive of the ground truth answer, so a model relying on it has learned the underlying task. We expect this feature to emerge late in training.
\end{enumerate}

\paragraph{Evaluating feature combinations.}
We fit a logistic regression model to predict the LM’s choice. The probability of selecting $e_1$ is given by $\hat{p} = \sigma\!\left( w^\top \phi(d, e_1, e_2) \right)$, where $w$ is learned by minimizing the standard logistic regression objective with L1 regularization.
The L1 penalty allows us to identify only the most predictive features. 
We use 5-fold cross-validation and select the regularization strength on the training folds. All features are standardized to zero mean and unit variance before fitting. We report the mean accuracy across the five folds.
To isolate individual strategies, we refit models using different subsets of features, including positional-only, frequency-only, and value-only feature sets, as well as models that include interaction features.

\paragraph{Metric for reporting feature ablations.} To characterize which strategies the LM uses at each point in training, we fit the regression features described above and report the \textbf{\emph{drop-one accuracy}}: the decrease in regression accuracy when a given feature is removed from the predictor set. 
This measures the importance of that feature in predicting LM behavior: a large drop-one accuracy for a feature indicates that the model's outputs are well-explained by that feature at that checkpoint.
We also define the \textbf{\emph{dominant phase}} at a checkpoint as the phase whose feature has the largest drop-one accuracy at that checkpoint.
\section{Learning Dynamics in Small LMs with Synthetic Data}
\label{sec:training}
To study how heuristics develop during training, we start with training small language models from scratch on mixtures of natural text from the OLMo-2 pretraining corpus and synthetic data from the entity comparison task. 
First, we vary the amount of synthetic task data relative to natural text and find that, with the right balance of data, three phases of learning emerge: a frequency heuristic first, then a position heuristic, and finally the correct algorithm (\Cref{sec:ratio}).
Then, we change the entity distributions in the training data from Zipfian to uniform and observe how this affects the development of these heuristics (\Cref{sec:distribution}).

\subsection{How the Ratio of Synthetic Data Shapes the Three Phases}
\label{sec:ratio}
Preliminary experiments demonstrated that the 163M-parameter LMs do not learn the comparison task when trained on only natural data, which is unsurprising at this scale. 
To help the model learn the behavior we want to study, we supplement natural text in the pretraining corpora with synthetic task-specific data. For the country area task, it follows the form \emph{``By total area, among the two countries A and B, the larger country is ''}. 
Under a fixed total data budget of 1.1B tokens, we vary the proportion of synthetic task supervision in 0\%, 0.01\%, 0.1\%, and 1\%. 
We generate synthetic data that matches the Zipfian distribution so that it better mirrors naturally occurring text. 
The fully natural setting serves as a control, which allows us to observe the effect of natural text on model behavior.

\begin{figure}[t]
  \centering
\includegraphics[width=0.85\linewidth]{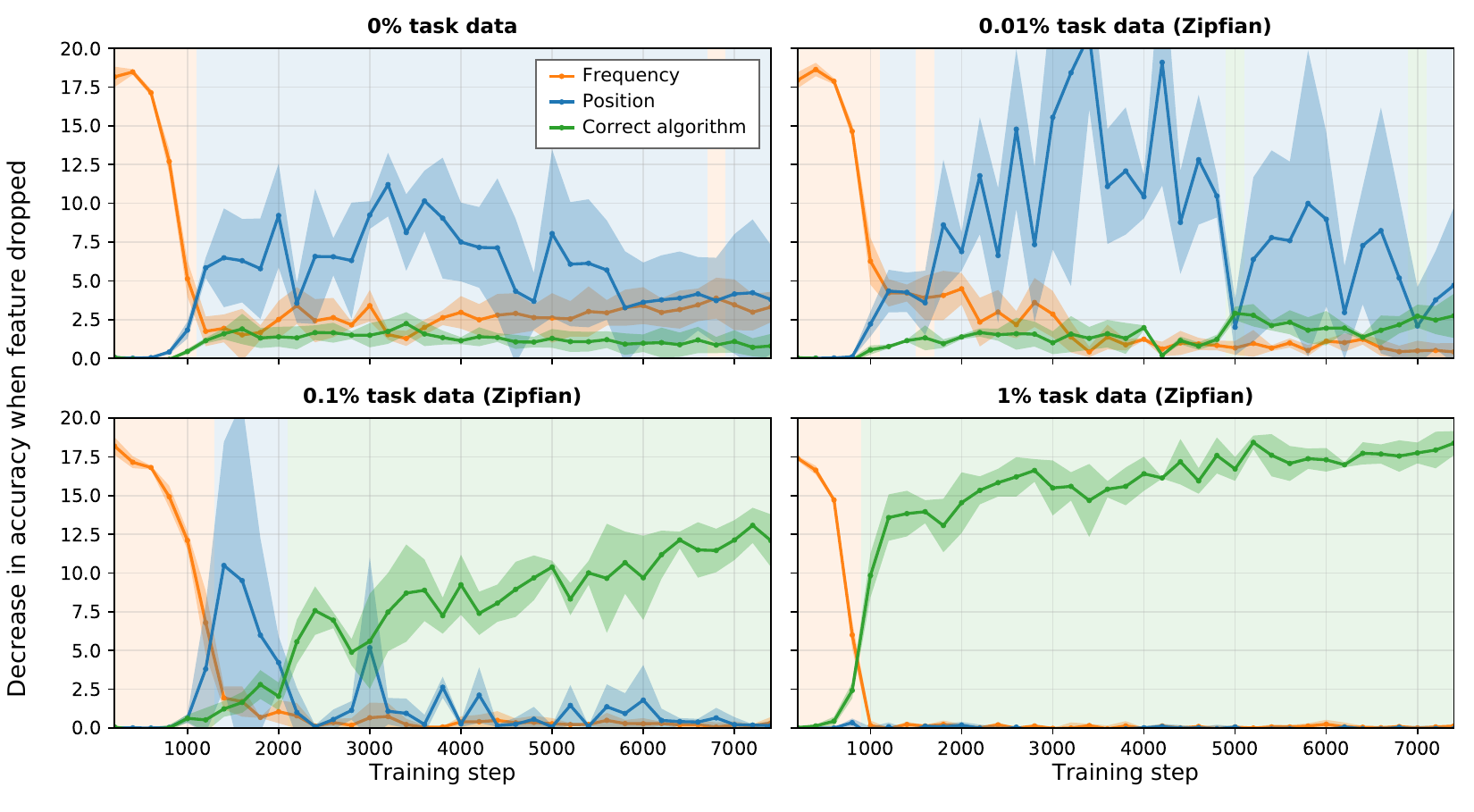}
    \caption{Learning trajectories across four ratios of synthetic task data to Zipfian natural text (0\%, 0.01\%, 0.1\%, 1\%). 
    Solid lines and dark shaded regions show mean and standard deviation across 3 training seeds.
    \textbf{The frequency phase is invariant to task supervision, but the correct algorithm phase only emerges above a critical density.} 
    Below this density (0\%, 0.01\%), the model never reaches the correct algorithm and instead defaults to positional shortcuts.}
    \label{fig:ratio}
\end{figure}

\paragraph{The proportion of task-specific synthetic data does not affect the frequency phase.}
\Cref{fig:ratio} shows that the frequency phase is essentially identical across all four conditions: it peaks at step 200 and decays by step 1K, regardless of synthetic ratio.
This motivates the experiments in the next section (\S \ref{sec:distribution}), where we show that the frequency phase depends on the entity frequency distribution of the training data, and that flattening that distribution weakens the phase.

\paragraph{The correct algorithm emerges only above a critical density of task specific synthetic data.}
The difference across conditions is whether the model ever reaches the correct algorithm. 
At 0\% and 0.01\% synthetic data, the correct-algorithm feature stays near zero throughout training. 
At 0.1\%, it emerges around 2.5K steps and rises steadily. 
When task data is increased to 1\%, it emerges within the first 1K steps and reaches a drop-one accuracy of $18$ by end of training. 
This corroborates existing work and suggests that there is a threshold: below a critical density of task-relevant signal, the model never finds learning the task loss-reducing enough to override existing shortcuts \citep{gu2025data,verma2025how}.

\paragraph{The position phase shrinks and eventually disappears as task-specific synthetic data increases.} When the correct algorithm phase does not emerge (0\% and 0.01\%), model behavior converges to a strong positional bias that persists throughout the rest of training. At 0.1\%, a shorter position phase still appears between the frequency and correct-algorithm phases. At 1\%, the position phase is absent as the model jumps directly from frequency to solving the task correctly.

\subsection{How Entity Frequency Distribution Drives the Heuristic Phases}\label{sec:distribution}

With 0.1\% of synthetic task data, we observe the three-phase learning dynamics emerge in training. 
However, the previous experiments use synthetic data with entity frequency distributions that match the Zipfian distribution of naturally occurring data.
Now, we separately manipulate the entity frequency distribution and change from Zipfian to uniform.
We run this experiment on three tasks, \textit{country area} and \textit{country GDP} comparison presented in the main paper, and \textit{person birth year} in Appendix~\ref{app:people_birthyear}. For each task, we train a separate model for each of the four natural vs. synthetic distribution combinations.
To construct the uniform setting, we sample sequences that contain country or people mentions to achieve roughly uniform mentions per entity.

\begin{figure}[t]
  \centering
  \includegraphics[width=\linewidth]{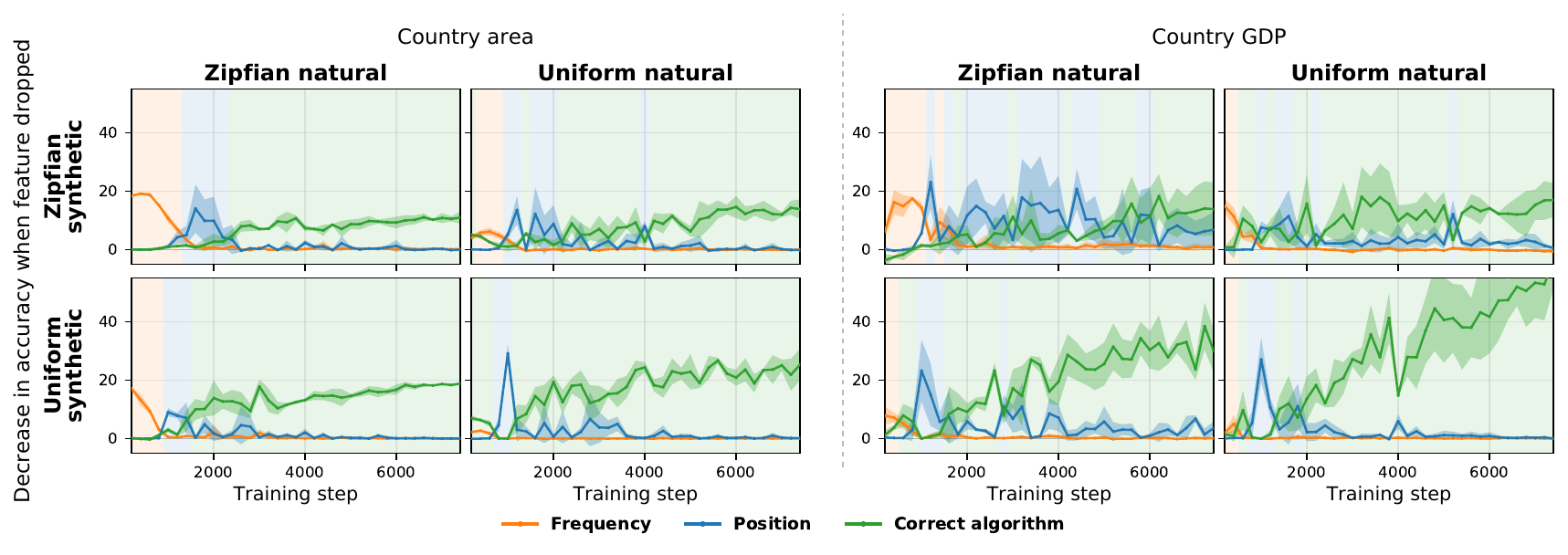}
  \caption{Learning trajectories for country area (left) and country GDP (right) across four combinations of natural text and synthetic data distributions (Zipfian $\times$ Uniform). \textbf{Both tasks show the same frequency, position, correct algorithm ordering. The frequency phase is strongest when natural text is Zipfian.} Solid lines and shaded regions show mean and standard deviation across 3 seeds.}
  \label{fig:data-distributions}
\end{figure}

\paragraph{The Zipfian distribution of natural text drives the initial frequency phase.}
\Cref{fig:data-distributions} shows the learning trajectories for country area
(left) and country GDP (right). 
In each block, the columns vary the entity distribution of the natural text (Zipfian or uniform), and the rows vary the distribution of the synthetic data.
When natural text is Zipfian (first column of each block), both tasks show a strong early frequency phase, regardless of the synthetic distribution.
When natural text is uniform, this phase is much weaker, and it nearly disappears when both natural text and synthetic data are uniform.
The correct algorithm shows the opposite pattern. When both distributions are uniform (bottom right of each block), it emerges earliest and reaches its largest drop-one accuracy. When both are Zipfian (top left), it is weakest.

\begin{figure}[h]
  \centering
  \includegraphics[width=0.78\linewidth]{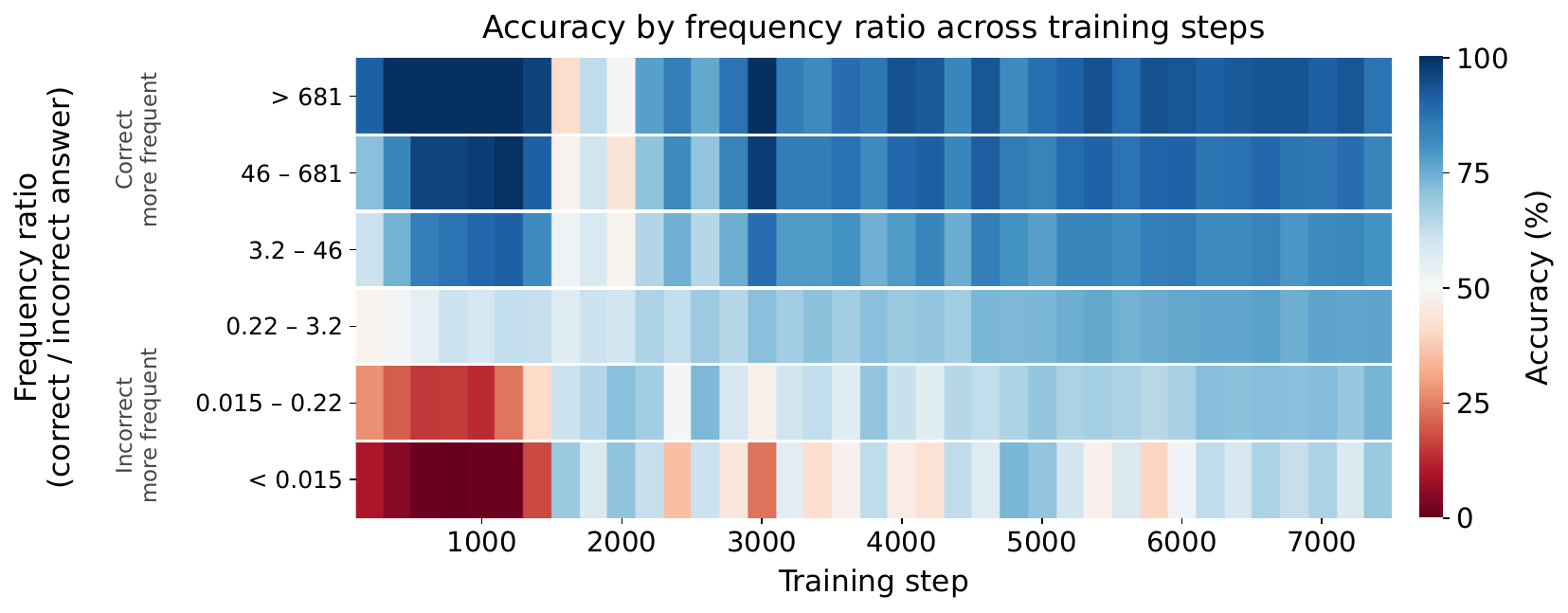}
  \caption{Accuracy by entity frequency ratio (correct / incorrect entity) across training steps under Zipfian data. Early in training, accuracy is high only when the correct answer is much more frequent, which corresponds to the frequency bias phase. Throughout training, accuracy is consistently higher when the correct entity is more frequent than when it is less frequent.}
  \label{fig:heatmap}
\end{figure}

\paragraph{The frequency heuristic is reduced through training but never fully eliminated.}
Grouping entity pairs by the ratio of the correct entity's frequency to that of the incorrect one, we find that this bias persists beyond the initial phase.
In Figure~\ref{fig:heatmap}, a ratio below 1 indicates that the correct answer is the less frequent entity; a ratio above 1 indicates that it is the more frequent one. 
Early in training, accuracy is high only when the correct answer is substantially more frequent than the incorrect one.
While this bias gradually diminishes, the model still maintains higher accuracy on pairs where the correct answer is more frequent in late training stages.

\section{Learning Dynamics in OLMo Pretraining}\label{sec:olmo}
The controlled experiments in \Cref{sec:training} establish that the Zipfian distribution of entities in text drives the frequency heuristic phase, and that training on such corpora follows a three-phase structure.
We now ask whether the same pattern appears in models trained at scale on natural text alone, without any synthetic task supervision.

\paragraph{Setup.} We focus on the OLMo-2 model family (1B, 7B, 13B, and 32B)~\citep{groeneveld-etal-2024-olmo}, which provides fully open-sourced training data and checkpoints. 
We apply the same regression analysis as described in \Cref{sec:methods} to track which strategy dominates over time. 
We evaluate 26 checkpoints per model, focusing primarily on stage 1 pretraining, with dense early coverage (every 1K steps from 1K to 10K) and sparser, log-spaced checkpoints afterward. 
To account for prompt variation, we evaluate each checkpoint with 3 prompt templates and report mean and standard deviation across them. 
The checkpoints and prompts used are presented in Appendix~\ref{app:olmo}.
We structure this analysis in two parts. First, we track the country area task across all four model sizes to observe how learning dynamics change with scale (\Cref{fig:olmo-trajectories}). Since the smaller models do not reliably learn the task, we then focus on the 32B model and extend the analysis to all nine comparison tasks, in both directions, to test how well the phase structure generalizes (\Cref{fig:olmo-32b-9tasks}).

\subsection{Phase Structure Across Model Scales}
\begin{figure}[t]
    \centering
\includegraphics[width=0.85\linewidth]{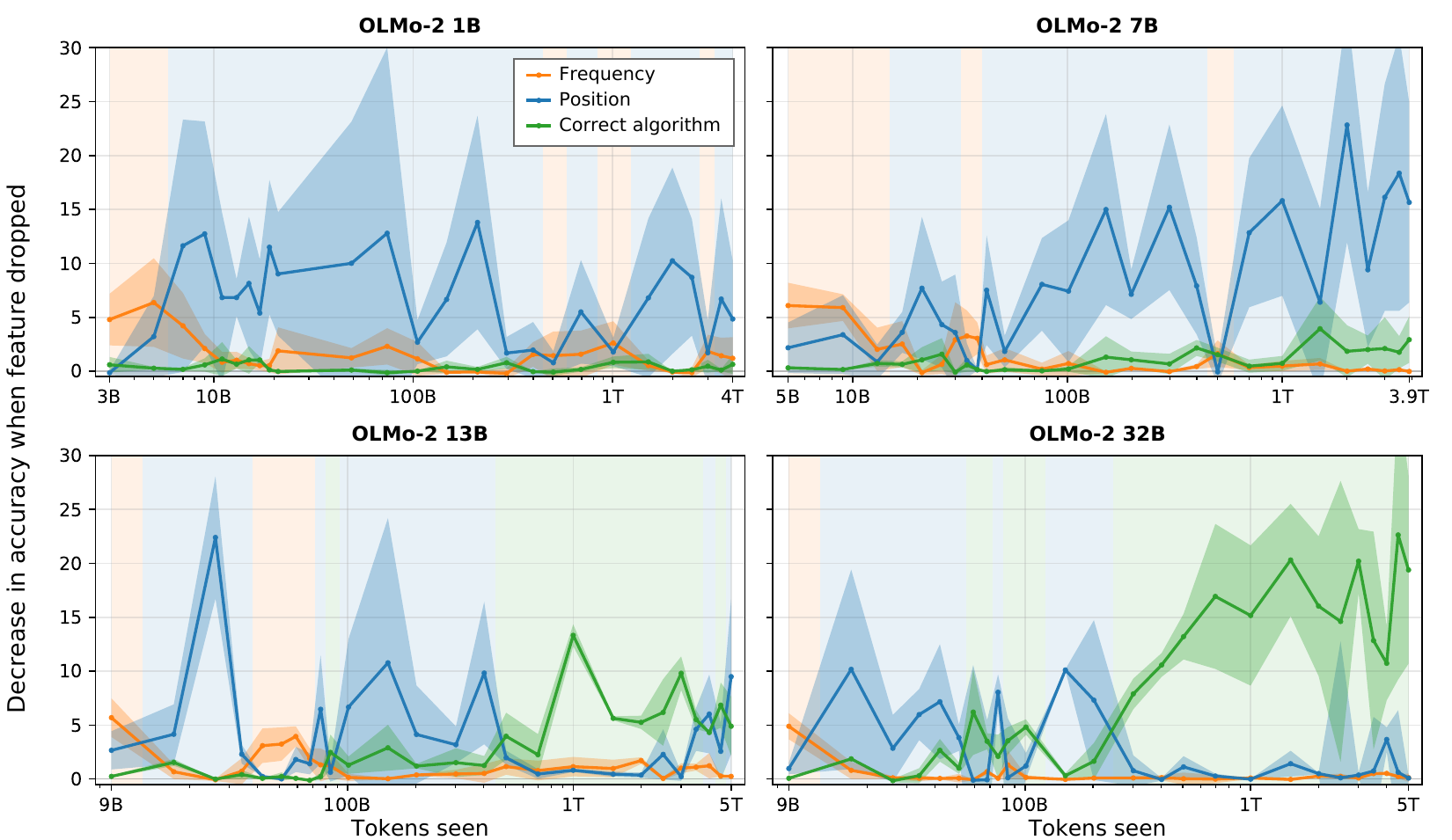}
    \caption{Feature importance across training checkpoints for OLMo-2 1B, 7B, 13B, and 32B. Solid lines and shaded regions show mean and standard deviation across 3 prompt variations per checkpoint. All models show frequency and position heuristics appearing in order, consistent with the phases observed in the controlled GPT-2 experiments. Smaller OLMo-2 models never learn the task.}
    \label{fig:olmo-trajectories}
\end{figure}

\paragraph{The frequency and position phases appear in a consistent order across model scales.}
In \Cref{fig:olmo-trajectories}, all four models exhibit the early heuristic phases observed in our controlled experiments, and they emerge in the order of frequency followed by position. 
The frequency heuristic appears briefly in the earliest checkpoint of all models.  
The 1B and 7B models then remain in a prolonged position heuristic phase without learning the task;
the 13B model transitions to the correct algorithm at around 1T tokens, while the 32B model does so earlier and achieves a higher peak.
We notice that the frequency phase has a lower peak than in the small LMs we pretrained, and hypothesize that this is because the first available checkpoint is already several billion tokens into training. A more pronounced frequency phase may have occurred before this point.
To check this, we rerun the official OLMo-2 1B stage-1 training from initialization for the first 1,000 steps (Appendix~\ref{app:olmo-early1000}). Frequency is already the dominant phase within the first 200 steps for most of our nine comparison tasks, supporting that the frequency phase begins before the first public checkpoint.

\paragraph{Only the 13B and 32B models learn the correct algorithm and exhibit all three phases.}
The correct algorithm phase emerges only in the two larger models, at 1T tokens for the 13B model and 400B tokens for 32B. 
In the two smaller models, the value feature stays near zero throughout training, meaning that they never learn the task. 
This parallels the threshold finding from the synthetic settings in \Cref{sec:ratio}: just as the correct algorithm fails to emerge below a critical density of task supervision in small-scale models, here it fails to emerge below a critical model size. 
Natural pretraining text contains relatively sparse task signal for entity comparison, and at 1B and 7B parameters the model lacks the capacity to learn the task.

\subsection{Generalization Across Nine Comparison Tasks}
To test how broadly the phase structure generalizes, we extend the analysis to all nine comparison tasks, in both the larger- and smaller-magnitude directions (18 task variants in total).
We focus on OLMo-2 32B, since early experiments show that the smaller models (especially 1B and 7B) do not reliably learn the tasks by the end of stage 1 pretraining.

\begin{figure}[ht]
    \centering
\includegraphics[width=0.9\linewidth]{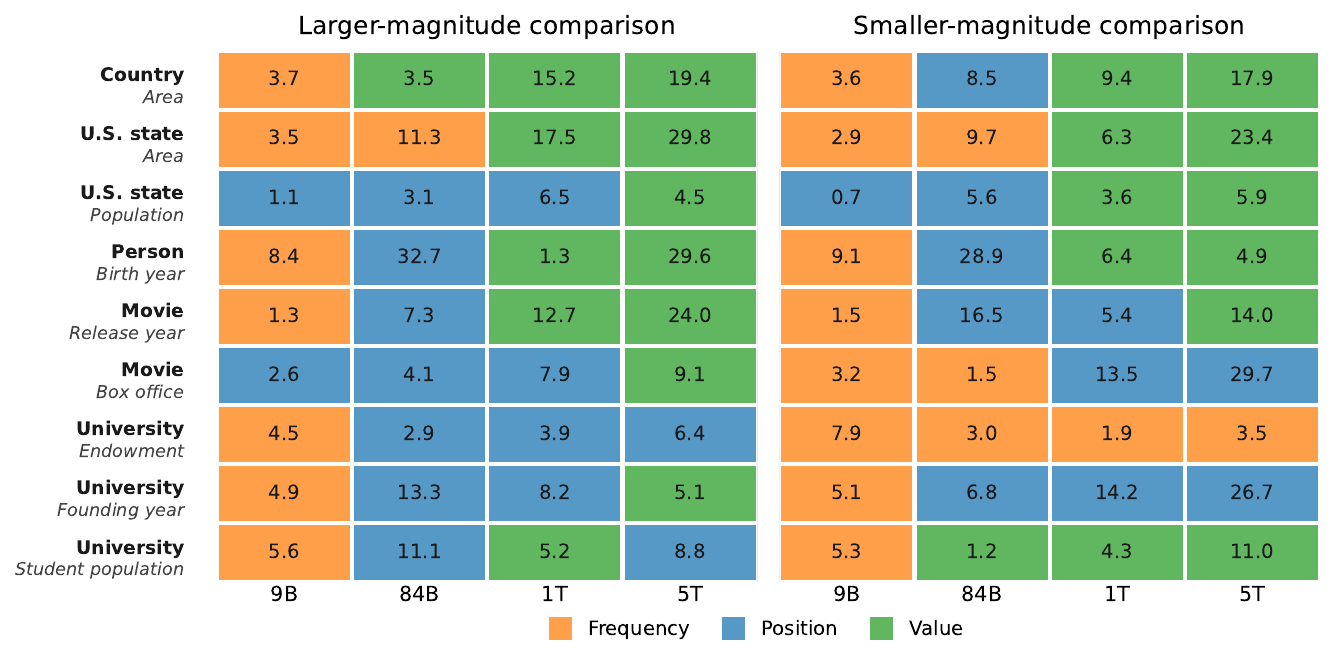}
    \caption{Dominant phase for OLMo-2 32B across all 18 task variants at four stage-1 checkpoints (9B, 84B, 1T, and 5T tokens). Each cell is colored by the feature with the largest drop-one accuracy at that checkpoint, and the number is that feature's drop-one accuracy (\%). Left: larger-magnitude comparisons; right: smaller-magnitude comparisons. \textbf{Frequency dominates the earliest checkpoint for most variants, and later phases largely follow the frequency--position--value structure.}}
    \label{fig:olmo-32b-9tasks}
\end{figure}

\paragraph{The frequency heuristic dominates at the earliest checkpoint for nearly every task.}
Figure~\ref{fig:olmo-32b-9tasks} shows the dominant phase at four selected checkpoints: the first available checkpoint (9B tokens seen), two intermediate checkpoints, and the final checkpoint (5000B tokens seen) at the end of stage 1 pretraining. Full results for all checkpoints are in Appendix~\ref{app:olmo-full}.
At the earliest checkpoint (9B tokens), frequency is the dominant phase in 15 of the 18 task variants.
This includes 8 of the 9 smaller-magnitude variants, where selecting the more frequent entity is typically incorrect. This suggests that the early phase reflects entity frequency statistics rather than partial task knowledge.

\paragraph{Phases can be skipped, but their ordering is largely preserved.}
Which phases appear varies by task, since for some tasks even the 32B model struggles to learn the correct algorithm. However, the relative order of the three phases is more consistent: across the training trajectory, whenever two phases both become dominant, they largely follow the frequency, position, value order.

\section{Discussion}

\paragraph{Implications on training.} 
Across scales from 163M to 32B parameters, models exhibit the same heuristic phases, and the 1B and 7B OLMo models never escape the positional heuristic. 
These heuristics appear to be a structural feature of how the task is learned, rather than an artifact of small or synthetic data. 
The key difference is whether the model successfully transitions out of these heuristics, not whether the heuristics emerge. 
In controlled experiments, the correct algorithm fails to emerge below a critical density of task supervision; in OLMo, it fails to emerge below a critical model scale on natural text alone.
This is in line with prior work sugesting that the correct strategy only becomes loss reducing once the model has sufficient task-relevant signal relative to its capacity.
Practically, our results indicate that the long tail in natural pretraining corpora may impose a real cost on downstream task acquisition. 
Flattening these heavy-tailed distributions in pretraining corpora can make learning more efficient.

\paragraph{Understanding of heuristics learning.}
We observe three phases in learning: first a frequency heuristic, then a position heuristic, and finally the correct algorithm. 
The consistency of this ordering across very different settings suggests that it reflects a structural property of the task: each phase corresponds to the easiest-to-learn predictive feature given the model’s current state. 
Frequency is directly available from token-level statistics, position depends on syntactic patterns in the data, and value-based entity comparison requires manipulating factual knowledge. 
Each subsequent strategy is likely more expensive to learn, but eventually becomes more loss reducing than its predecessor. 

\paragraph{Limitations.} 
We focus our experiments on the task of entity comparison, and the observed phase structure may not generalize to tasks with different structures.
Our controlled experiments are conducted at the 163M scale in order to efficiently vary data composition and perform model pretraining under feasible compute budgets, which would otherwise be infeasible for large-scale pretraining.
Although these intervention experiments are small-scale, we show that the toy setting can nevertheless provide insight into the training dynamics of much larger models, which is itself an interesting finding. 
While our OLMo-2 analysis suggests that the same phases emerge at much larger scales, we cannot establish causal evidence for the role of data at scale.

Lastly, our experiments that modify pretraining data statistics (e.g., changing the country distribution from Zipfian to uniform) require re-selecting sequences containing the target entities. For example, this may involve excluding sequences heavily focused on United States politics while selecting sequences about tropical fruits in Madagascar.
This process could introduce domain shifts and other confounding factors that we do not account for. 
There is currently no straightforward solution to this issue, and it remains an active open research problem \citep{nadkarni2025rewriting}.

\paragraph{Future work.}
We would like to understand the underlying mechanisms for each heuristic and more fully trace the origins of the three phases in this order.
For example, whether the positional bias relies on induction-head-style in-context retrieval and if it relates to when induction heads are formed, and whether there exists latent representations that track entity frequency.
\section{Related Work}\label{sec:related-work}

\paragraph{Heuristics and different task strategies.}
LMs often rely on heuristics rather than the intended algorithms for solving tasks. 
Prior work has identified shortcuts in natural language inference~\citep{mccoy2019right}, arithmetic reasoning~\citep{razeghi2022impact, nikankinarithmetic}, multi-hop reasoning~\citep{yang-etal-2025-large}, and factual recall, where entity popularity sharply predicts whether a model answers correctly~\citep{mallen-etal-2023-trust, kandpal2023large}. 
Most closely related to our setting, \citet{lehmann2026knowing} study entity comparison in fully trained LLMs and identify three heuristic biases that influence predictions. 
A complementary view characterizes LM behavior as a mixture of strategies rather than a single mechanism~\citep{wutransformers,gur2025mixing, pislar2025combining, todd2025context}. These analyses operate on static checkpoints; we instead study the learning trajectory itself, asking when each heuristic emerges during training and what data properties drive its emergence.

\paragraph{Pretraining data.}
Pretraining data affects model behavior, performance, and internal representations~\citep{elazar2022measuring,mccoy2023embers, shaib2024detection, merullolinear, ruis2025procedural,verma2025how}. 
\citet{wang2025generalization} measure the correlation between LLM output probabilities and the frequency of task-relevant n-gram pairs in pretraining data, while other work has traced model biases and failures to asymmetries in entity frequency~\citep{razeghi2022impact, he2025supposedly, kandpal2023large, mallen-etal-2023-trust}. 
To make causal claims about data effects, prior work has intervened directly on training data composition~\citep{wei2021frequency, chan2022data, allen2024physics}; \citet{nadkarni2025rewriting} provide a recipe for document-level interventions, and \citet{longpre-etal-2024-pretrainers} systematically study effects of data age, domain coverage, and quality. 
Our work builds on this interventional methodology and identifies entity distribution as a causal driver of which heuristic phases emerge during training.

\paragraph{Phases and training dynamics.}
Capabilities are often acquired during pretraining through abrupt transitions or distinct phases that build on earlier representations~\citep{saphra-lopez-2019-understanding, chen2024sudden, wutransformers}. 
A canonical case study is the formation of induction heads, which coincides with a sharp drop in loss and the emergence of in-context learning~\citep{olsson2022context, singh2024needs}. 
Other work tracks how the internal mechanisms behind a capability change over training: the attention heads implementing a circuit can be replaced over time while the algorithm they implement stays the same~\citep{tigges2024llm}, and components can shift from general-purpose to task-specific as a model gets better at predicting an answer~\citep{hakimi-etal-2025-time}.
Closest to our setting, \citet{zucchet2025how} also find that factual recall emerges in distinct phases, with a plateau before precise recall that shortens under more balanced training data; \citet{chang2024how} similarly study how factual knowledge is acquired and forgotten over the course of pretraining.
\citet{ravfogel2026emergence} trace the emergence of a linear truth direction, and \citet{gu2025data} study phase transitions as a function of data mixing ratio. 
Our work extends this perspective to natural pretraining data statistics.
\section{Conclusion}
We identify three phases of learning in entity comparison and show through controlled experiments that pretraining data composition causally determines which phases emerge and when.
The same phase structure appears in OLMo-2 models trained on natural text at scales up to 32B parameters, suggesting these dynamics are a general feature of how LMs learn this task.
Our results highlight pretraining data statistics as both a driver of heuristic learning and a controllable lever for studying it.

\begin{ack}
We thank Junyi Tao for helping with early experiments. YSC thanks members of the LUNAR lab and BATS lab at Brown University for helpful discussions at early stages of the project. This research was supported in part by a grant from Open Philanthropy and by the Israeli Science Foundation (Grant Number 4230/26).
\end{ack}

\bibliography{citations}
\bibliographystyle{plainnat}

\newpage
\newpage
\appendix
\section{Tasks and Prompt Templates}\label{app:tasks}

\Cref{tab:all-task-prompts} lists all nine comparison tasks we study, which spans five entity types, with one example prompt per task (the larger magnitude variant, with concrete entities filled in for illustration). 
Each task also has a smaller magnitude variant, evaluated identically but with the comparator word(s) inverted (e.g., \textit{``the larger country''} to \textit{``the smaller country''}; \textit{``born more recently''} to \textit{``born earlier''}).
This gives 18 task variants in total. All prompts follow the same style of natural text continuation.

\begin{table}[htbp]
\centering
\small
\caption{The nine comparison tasks with one example prompt per task.}
\label{tab:all-task-prompts}
\begin{tabular}{ll p{0.65\linewidth}}
\toprule
\textbf{Entity} & \textbf{Domain} & \textbf{Example prompt} \\
\midrule
Country & Area &
By total area, among the two countries France and Brazil, the larger country is \\[4pt]
U.S.\ state & Area &
By total area, among the two U.S.\ states Texas and Ohio, the larger state is \\[4pt]
U.S.\ state & Population &
By population, among the two U.S.\ states California and Vermont, the more populous state is \\[4pt]
Person & Birth year &
Between the two people Socrates and Barack Obama, the one born more recently is \\[4pt]
Movie & Release year &
Between the two films Casablanca and Inception, the one released more recently is \\[4pt]
Movie & Box office &
By worldwide box office gross, among the two films The Dark Knight and Pulp Fiction, the higher-grossing film is \\[4pt]
University & Endowment &
By financial endowment, among the two universities Harvard University and Rice University, the wealthier university is \\[4pt]
University & Founding year &
Between the two universities Harvard University and Stanford University, the one founded more recently is \\[4pt]
University & Student population &
By number of enrolled students, among the two universities The Ohio State University and Williams College, the larger university is \\
\bottomrule
\end{tabular}
\end{table}
\FloatBarrier

\section{Feature Ablation Experiments}\label{app:feature-ablation}

\subsection{List of Evaluated Features}\label{app:feature-list}
To identify the strategies that LMs use for the entity comparison task, we evaluate a wide range of features in initial experiments.
For each input triple $(d, e_1, e_2)$, we construct a feature vector $\phi(d,e_1,e_2)$ to capture relevant features that correspond to possible heuristics. We consider four families of features:
\begin{enumerate}
    \item \textbf{Positional bias feature}: binary indicators of left/right position, independent of entity semantics or values (e.g., whether $e_1$ is favored by virtue of being mentioned first). This is the intercept in the logistic regression model. 
    \item \textbf{Frequency-based features}: entity occurrence and co-occurrence counts in the pretraining corpus (e.g., sum of the occurrences of $e_1$ and $e_2$). 
    \item \textbf{Value-based features}: variants of the numerical values $v_{d,e_1}$ and $v_{d,e_2}$, including absolute values, log-scaled values, differences, and ratios.
    \item \textbf{Related-domain features}: values of numerical properties other than $d$ that may co-vary with $v_{d,e}$ (e.g., population or GDP when $d$ is area), intended to test whether models rely on these correlated proxies.
\end{enumerate}

For the frequency and domain values, we include normalized variants (e.g., z-scores, distance-to-mean) to test if the model responds to relative magnitude rather than absolute values. 
The full list of features and their corresponding definitions is provided in \Cref{tab:feature-definitions}.

\begin{table}[htbp]
\centering
\small
\caption{Features evaluated for modeling the LM behavior. Examples are illustrative values for the country area task with $e_1=$\,France ($v_{d,e_1}=551{,}500$\,km$^2$, $f_{e_1}=211{,}562{,}884$) and $e_2=$\,Germany ($v_{d,e_2}=357{,}581$\,km$^2$, $f_{e_2}=216{,}526{,}676$).}
\label{tab:feature-definitions}
\begin{tabular}{lll}
\toprule
\textbf{Name} & \textbf{Definition} & \textbf{Example} \\
\midrule
\multicolumn{3}{l}{\textit{Value-based features}} \\
\midrule
\texttt{value\_left} & $v_{d,e_1}$ & $551{,}500$ \\
\texttt{value\_right} & $v_{d,e_2}$ & $357{,}581$ \\
\texttt{value\_left\_log} & $\log_{10}(v_{d,e_1})$ & $5.74$ \\
\texttt{value\_right\_log} & $\log_{10}(v_{d,e_2})$ & $5.55$ \\
\texttt{value\_ratio} & $\log_{10}\!\left(v_{d,e_1} / v_{d,e_2}\right)$ & $0.19$ \\
\texttt{value\_left\_from\_threshold}  & $\left(v_{d,e_1} - \mu_v\right) / \mu_v$ & $-0.08$ \\
\texttt{value\_right\_from\_threshold} & $\left(v_{d,e_2} - \mu_v\right) / \mu_v$ & $-0.40$ \\
\texttt{value\_zscore\_left}  & $\left(\log_{10} v_{d,e_1} - \mu_{\log v}\right) / \sigma_{\log v}$ & $0.89$ \\
\texttt{value\_zscore\_right} & $\left(\log_{10} v_{d,e_2} - \mu_{\log v}\right) / \sigma_{\log v}$ & $0.75$ \\

\midrule
\multicolumn{3}{l}{\textit{Frequency-based features}} \\
\midrule
\texttt{freq\_left} & $f_{e_1}$ & $211{,}562{,}884$ \\
\texttt{freq\_right} & $f_{e_2}$ & $216{,}526{,}676$ \\
\texttt{freq\_left\_log} & $\log_{10}(f_{e_1})$ & $8.33$ \\
\texttt{freq\_right\_log} & $\log_{10}(f_{e_2})$ & $8.34$ \\
\texttt{freq\_sum} & $f_{e_1} + f_{e_2}$ & $428{,}089{,}560$ \\
\texttt{freq\_sum\_log} & $\log_{10}(f_{e_1}) + \log_{10}(f_{e_2})$ & $16.66$ \\
\texttt{freq\_diff} & $f_{e_1} - f_{e_2}$ & $-4{,}963{,}792$ \\
\texttt{log\_freq\_ratio} & $\log_{10}(f_{e_1}) - \log_{10}(f_{e_2})$ & $-0.01$ \\
\texttt{freq\_zscore\_left}  & $\left(\log_{10} f_{e_1} - \mu_{\log f}\right) / \sigma_{\log f}$ & $1.92$ \\
\texttt{freq\_zscore\_right} & $\left(\log_{10} f_{e_2} - \mu_{\log f}\right) / \sigma_{\log f}$ & $1.93$ \\
\texttt{cooccurrence\_log} & $\log_{10}\!\left(1 + c(e_1, e_2)\right)$ & $3.54$ \\
\texttt{cooccurrence\_larger\_string} & Directional pair count aligned to $e_1/e_2$ & $1{,}820$ \\

\midrule
\multicolumn{3}{l}{\textit{Related-domain features (computed for $a \in \{\textnormal{gdp}, \textnormal{population}\}$)}} \\
\midrule
\texttt{$a$\_left\_log}  & $\log_{10}(a_{e_1})$ & $\textnormal{gdp}_L\!=\!12.50$ \\
\texttt{$a$\_right\_log} & $\log_{10}(a_{e_2})$ & $\textnormal{gdp}_R\!=\!12.67$ \\
\texttt{$a$\_ratio}      & $\log_{10}(a_{e_1}) - \log_{10}(a_{e_2})$ & $\textnormal{gdp}\!=\!-0.17$ \\
\texttt{$a$\_sum\_log}   & $\log_{10}(a_{e_1}) + \log_{10}(a_{e_2})$ & $\textnormal{gdp}\!=\!25.17$ \\
\texttt{$a$\_zscore\_left}  & $\left(\log_{10} a_{e_1} - \mu_{\log a}\right) / \sigma_{\log a}$ & $\textnormal{gdp}\!=\!0.41$ \\
\texttt{$a$\_zscore\_right} & $\left(\log_{10} a_{e_2} - \mu_{\log a}\right) / \sigma_{\log a}$ & $\textnormal{gdp}\!=\!0.55$ \\
\bottomrule
\end{tabular}
\end{table}

\subsection{Validating Positional Bias Feature}\label{app:position_bias}
Because every entity pair is evaluated in both prompt orders (\Cref{sec:methods}), we can classify each unordered pair into one of four categories: the model chooses whichever entity appears \textit{first}, regardless of which entity that is; the model chooses whichever entity appears \textit{second}; the model chooses the \textit{correct} entity in both orderings; or the model chooses the \textit{incorrect} entity in both orderings. 
A strong position drop-one accuracy should correspond to a large share of pairs falling into the first two categories, while a stable entity-specific bias (e.g., from frequency, value, or tokenization) would instead fall into the last two categories. 
\Cref{tab:position-pair-symmetry} reports this breakdown for task and checkpoint combinations that have strong position drop-one accuracy, together with country area at a late checkpoint where the position bias is weak, as a contrast.

\begin{table}[htbp]
\centering
\small
\caption{Breakdown of entity pairs at representative task/checkpoint combinations, OLMo-2 32B (larger magnitude comparison). Position drop acc. is the position feature's drop-one accuracy (\%) at that checkpoint.}
\label{tab:position-pair-symmetry}
\resizebox{\textwidth}{!}{%
\begin{tabular}{lrrrrrr}
\toprule
\textbf{Task} & \textbf{Ckpt} & \textbf{Position drop acc.} & \textbf{First position} & \textbf{Second position} & \textbf{Always correct} & \textbf{Always incorrect} \\
\midrule
People birth year & 84B & 32.7 & 0.1 & 72.5 & 19.5 & 7.9 \\
Movie box office & 1T & 7.9 & 42.3 & 0.3 & 43.2 & 14.2 \\
University founding year & 1T & 8.2 & 0.9 & 34.1 & 34.6 & 30.4 \\
University student population & 5T & 8.8 & 37.2 & 0.3 & 53.2 & 9.3 \\
Country area & 5T & 0.1 & 8.2 & 4.6 & 78.1 & 9.1 \\
\bottomrule
\end{tabular}%
}
\end{table}

When the position feature has high drop-one accuracy, most unordered pairs fall into the first position or second position categories: the model's chosen entity changes when the prompt order is reversed, which confirms that the intercept is capturing order dependence rather than a stable entity-specific preference. 
In the last row where the position feature is weak, the large majority of pairs fall into the always correct category, meaning that the a model that has moved past the position heuristic.

\subsection{Feature Importance in Small LMs}\label{app:feature-importance-small}
This section presents full results from the controlled pretraining experiments in \Cref{sec:distribution}.
In addition to the the drop-one accuracy plots, we fit a logistic regression on every non-empty subset of the three main features (frequency, position, value) at every checkpoint, and report the regression's prediction accuracy on the LM's pairwise output.

Each figure has six panels, one per non-trivial subset of \{P, F, V\}: \{P\}, \{F\}, \{V\}, \{P{+}F\}, \{P{+}V\}, \{F{+}V\}. 
In every panel, the colored line is the prediction accuracy when only that subset is used as predictors, and the black line is the prediction accuracy of the full P{+}F{+}V regression on the same checkpoint, reproduced as a baseline. 
A subset whose curve sits on top of the full P{+}F{+}V baseline at a given checkpoint already captures all the LM's behavior at that point in training; a subset whose curve sits well below the baseline is missing a feature the LM is using.

\begin{figure}[htbp]
  \centering
  \includegraphics[width=\linewidth]{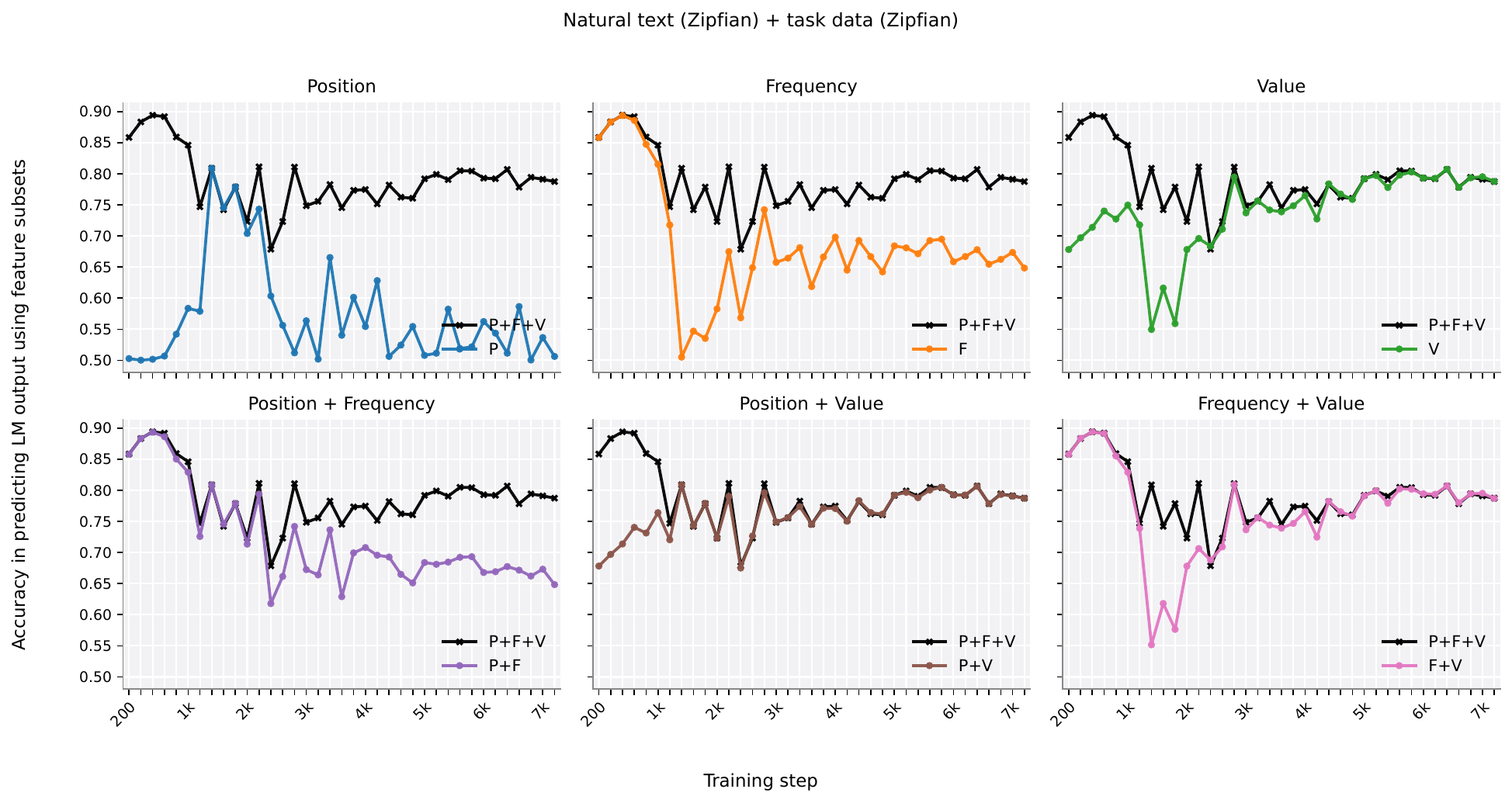}
  \caption{Prediction accuracy of feature subsets for the small LM trained on Zipfian natural text mixed with 0.1\% Zipfian synthetic data. Each panel compares one feature subset (colored line) against the full P{+}F{+}V regression (black).}
  \label{fig:subset-zipnat-zipsynth}
\end{figure}

\begin{figure}[htbp]
  \centering
  \includegraphics[width=\linewidth]{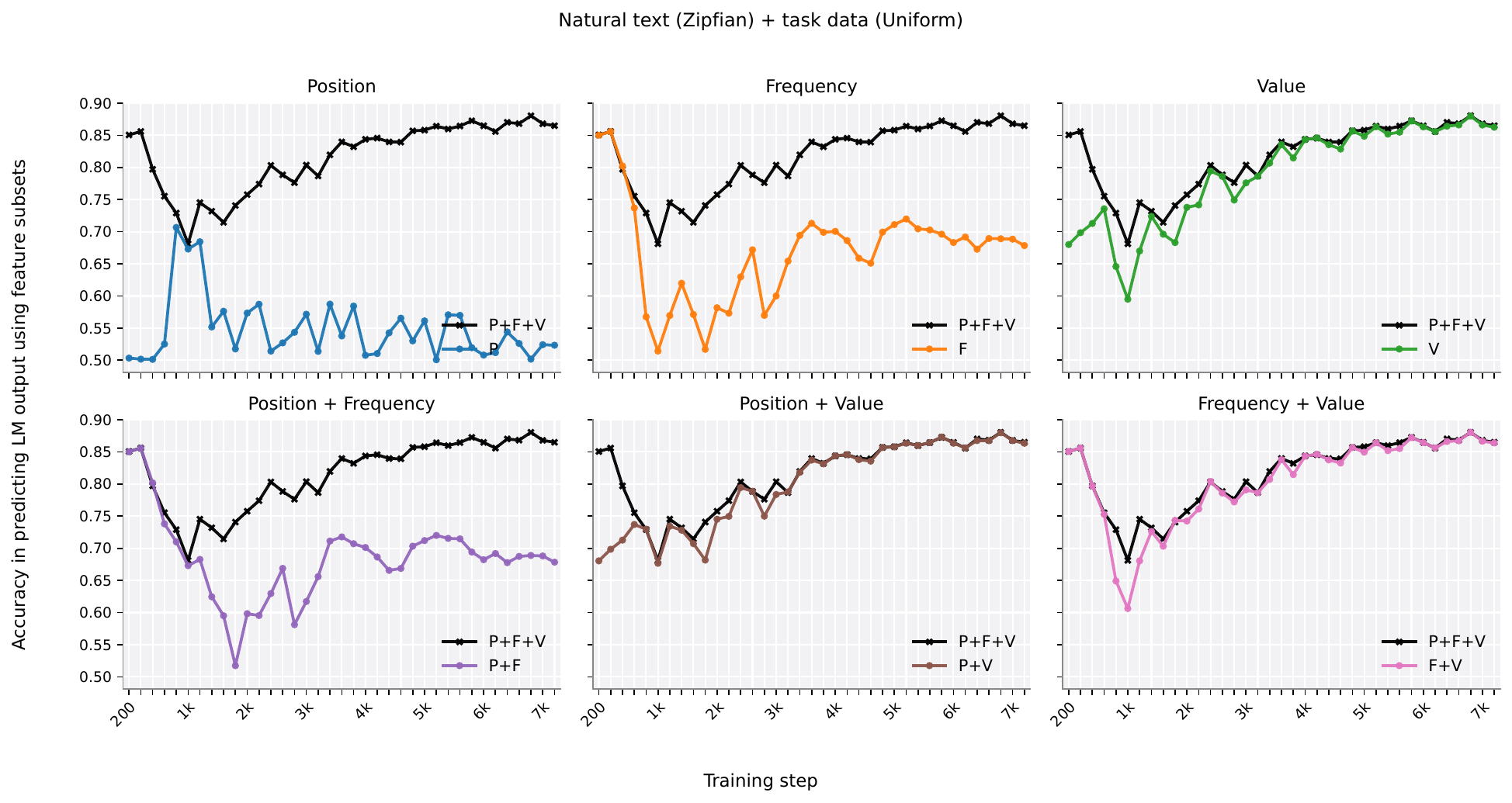}
  \caption{Prediction accuracy of feature subsets for the small LM trained on Zipfian natural text mixed with 0.1\% uniform synthetic data.}
  \label{fig:subset-zipnat-unisynth}
\end{figure}

\begin{figure}[htbp]
  \centering
  \includegraphics[width=\linewidth]{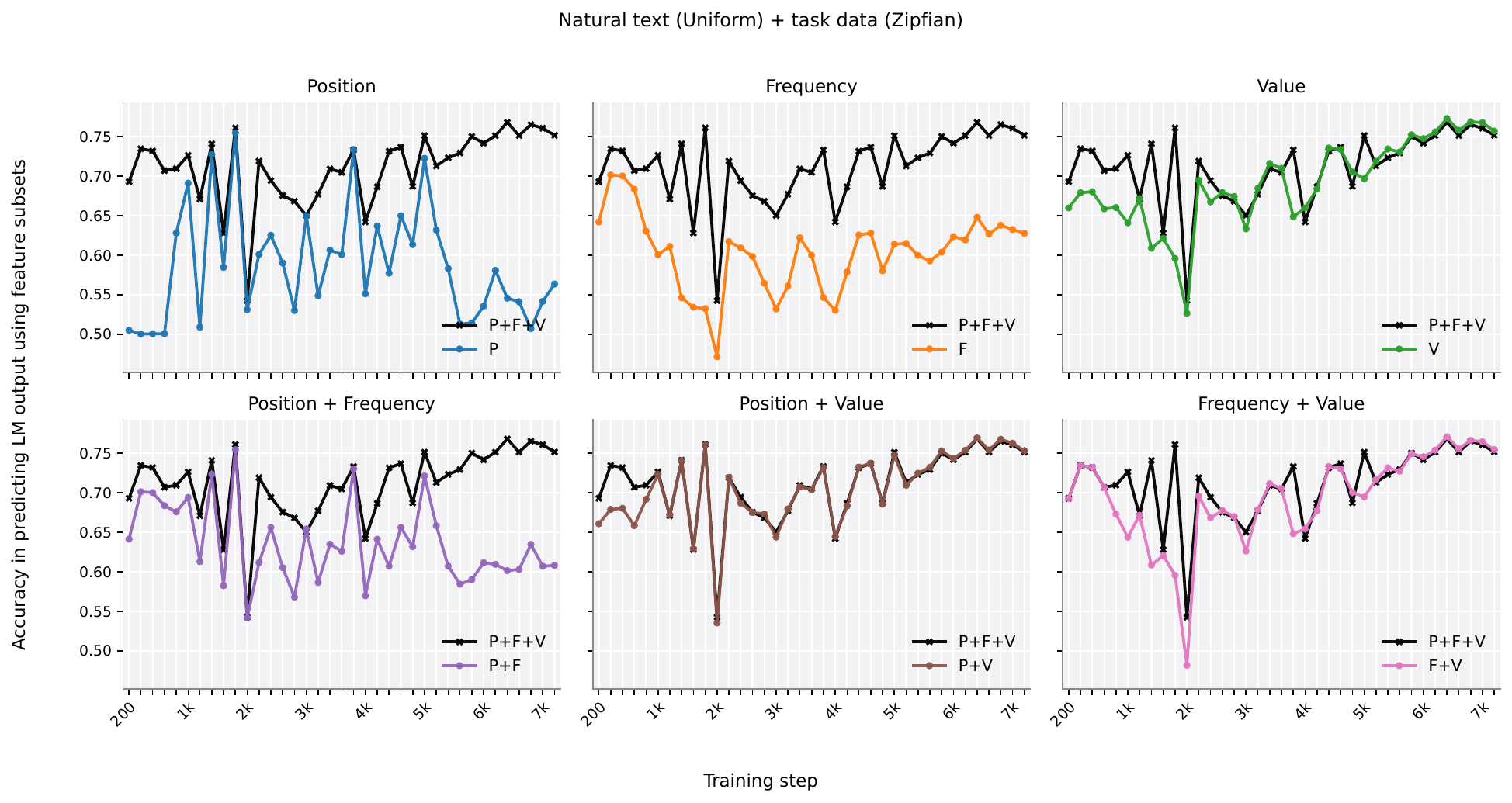}
  \caption{Prediction accuracy of feature subsets for the small LM trained on uniform natural text mixed with 0.1\% Zipfian synthetic data.}
  \label{fig:subset-unifnat-zipsynth}
\end{figure}

\begin{figure}[htbp]
  \centering
  \includegraphics[width=\linewidth]{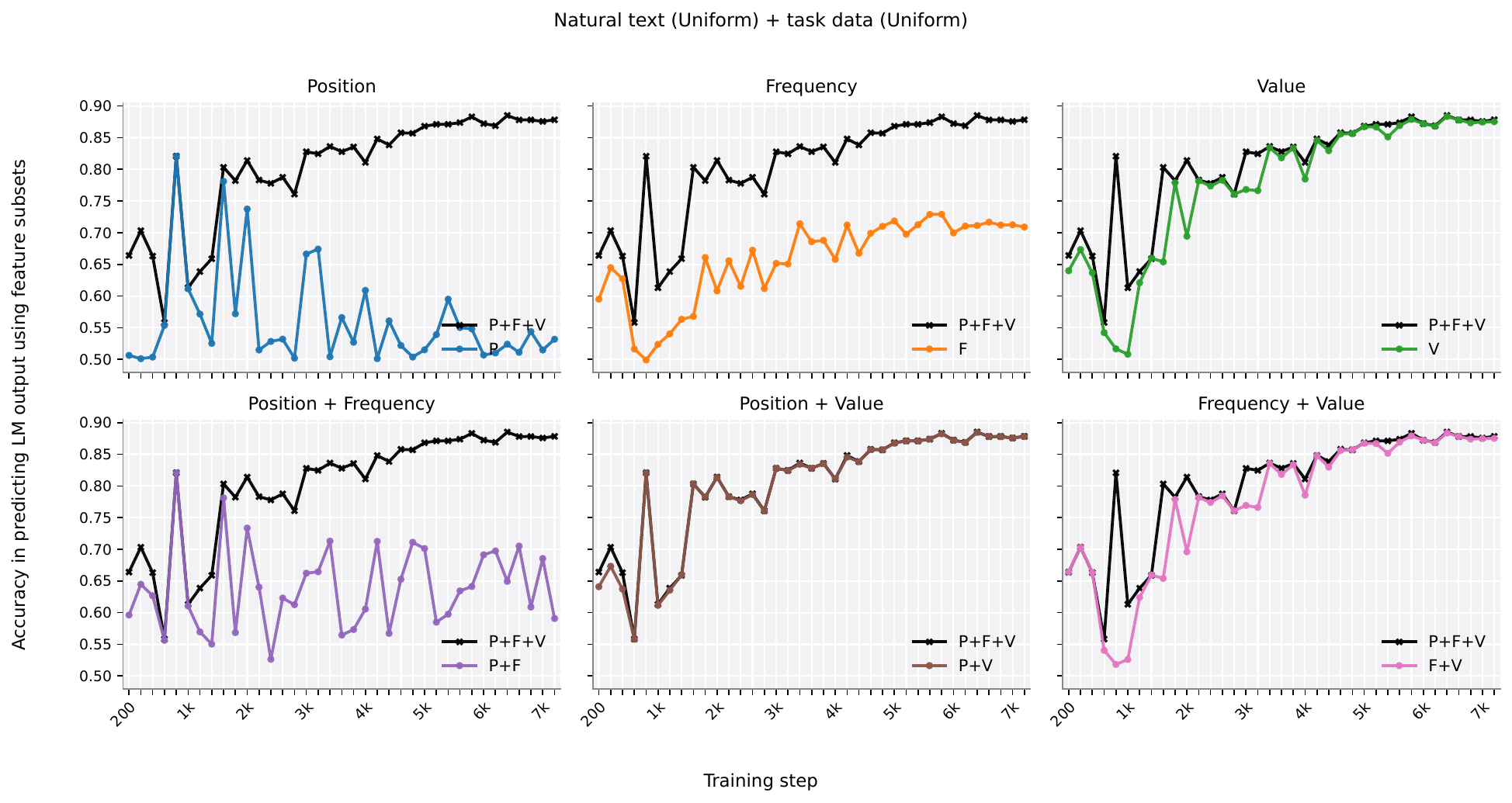}
  \caption{Prediction accuracy of feature subsets for the small LM trained on uniform natural text mixed with 0.1\% uniform synthetic data.}
  \label{fig:subset-unifnat-unisynth}
\end{figure}

\FloatBarrier
\clearpage

\newpage
\subsection{Feature Importance in OLMo Models}\label{app:feature-importance-olmo}
\FloatBarrier

We repeat the feature subset regression analysis on the OLMo-2 1B, 7B, 13B, 32B checkpoints listed in \Cref{tab:olmo_checkpoints}, which are pretrained on natural text alone with no synthetic task supervision. 
As in \Cref{app:feature-importance-small}, each figure has six panels, one per non-trivial subset of \{P, F, V\}. The colored line in each panel is the prediction accuracy of the regression on that subset; the black line is the prediction accuracy of the full P{+}F{+}V regression baseline on the same checkpoint.

\begin{figure}[ht]
  \centering
  \includegraphics[width=\linewidth]{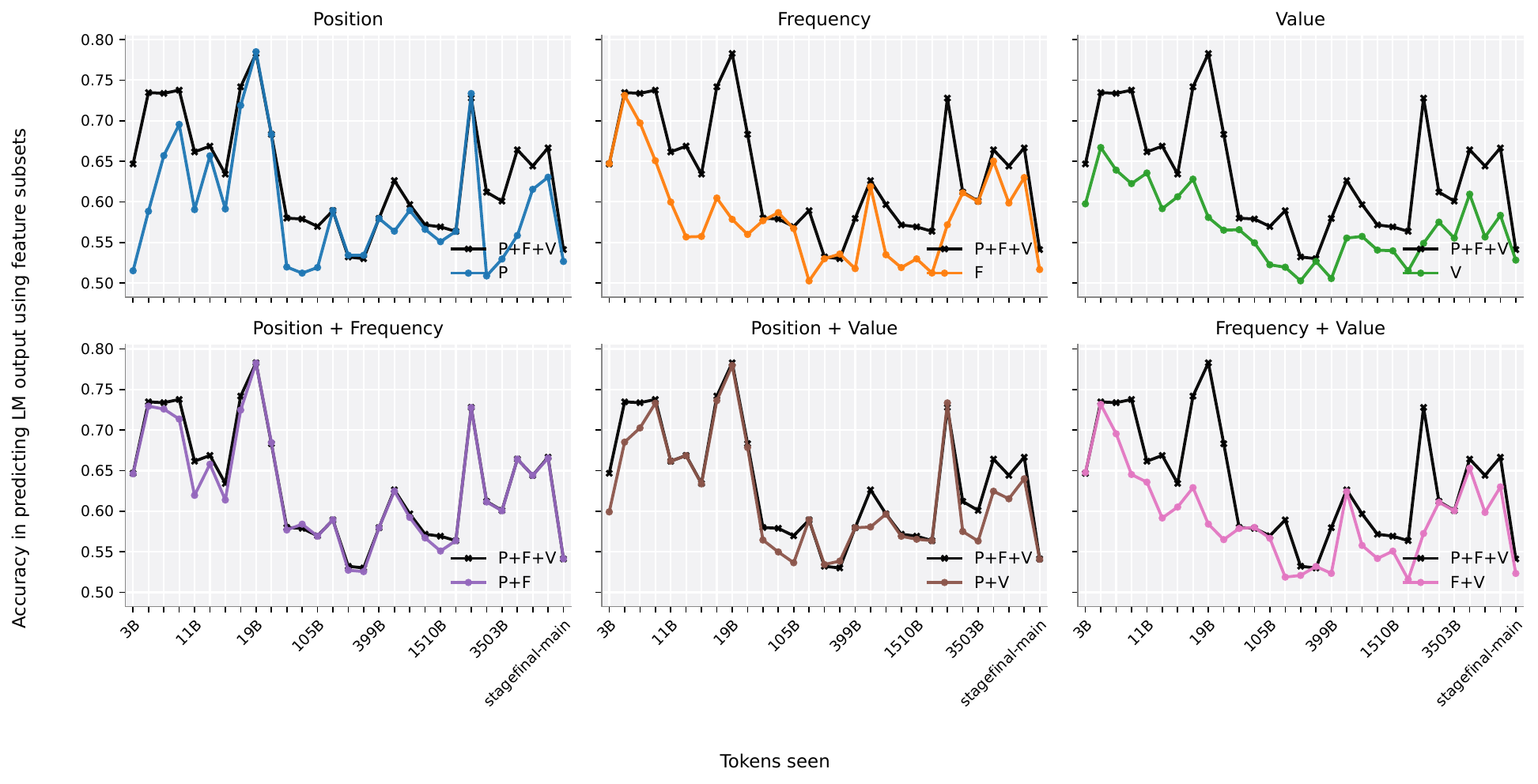}
  \caption{Prediction accuracy of feature subsets for OLMo-2 1B training checkpoints. Each panel compares one feature subset (colored line) against the full P{+}F{+}V regression (black).}
  \label{fig:subset-olmo-1b}
\end{figure}

\begin{figure}[htbp]
  \centering
  \includegraphics[width=\linewidth]{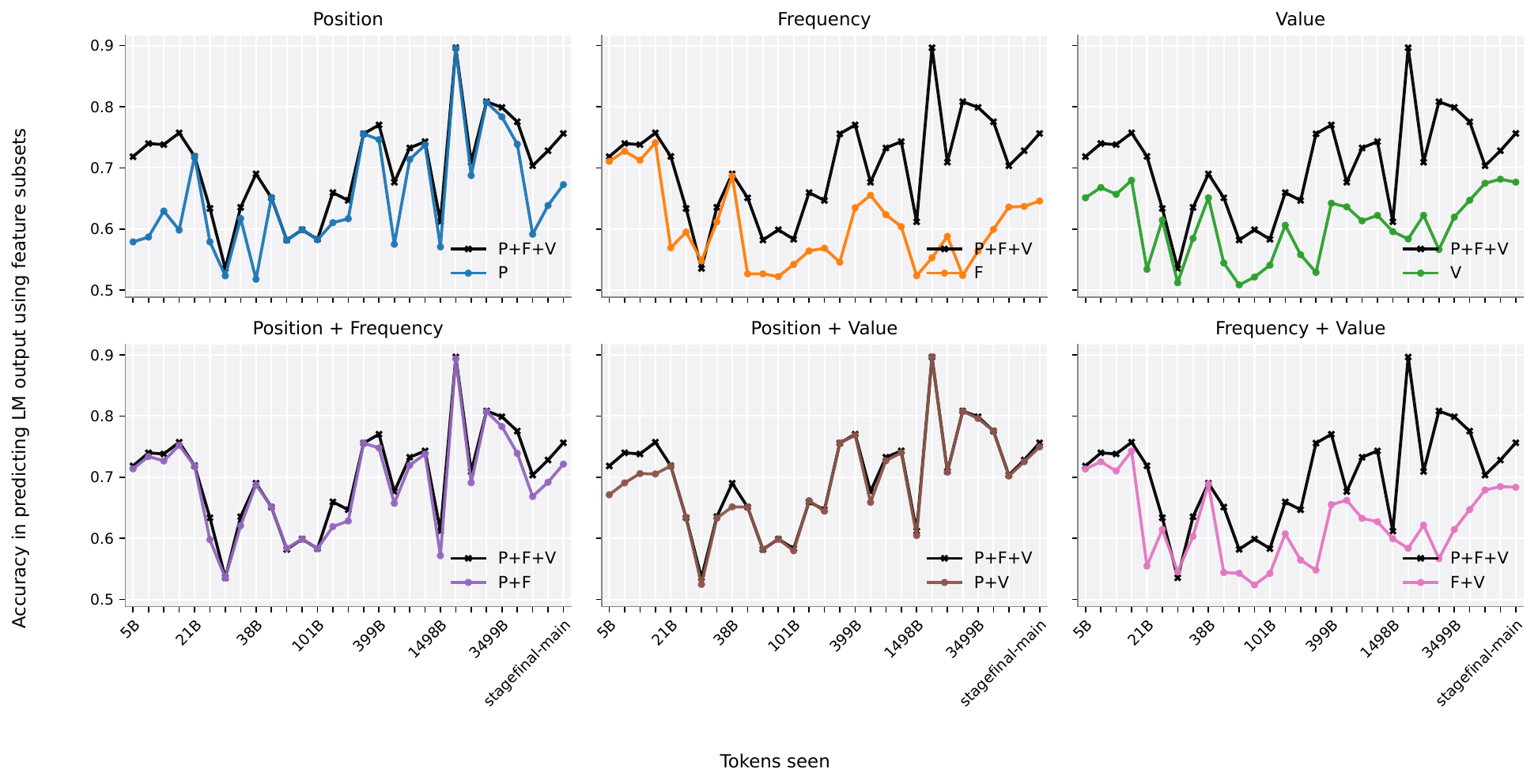}
  \caption{Prediction accuracy of feature subsets for OLMo-2 7B training checkpoints.}
  \label{fig:subset-olmo-7b}
\end{figure}

\begin{figure}[htbp]
  \centering
  \includegraphics[width=\linewidth]{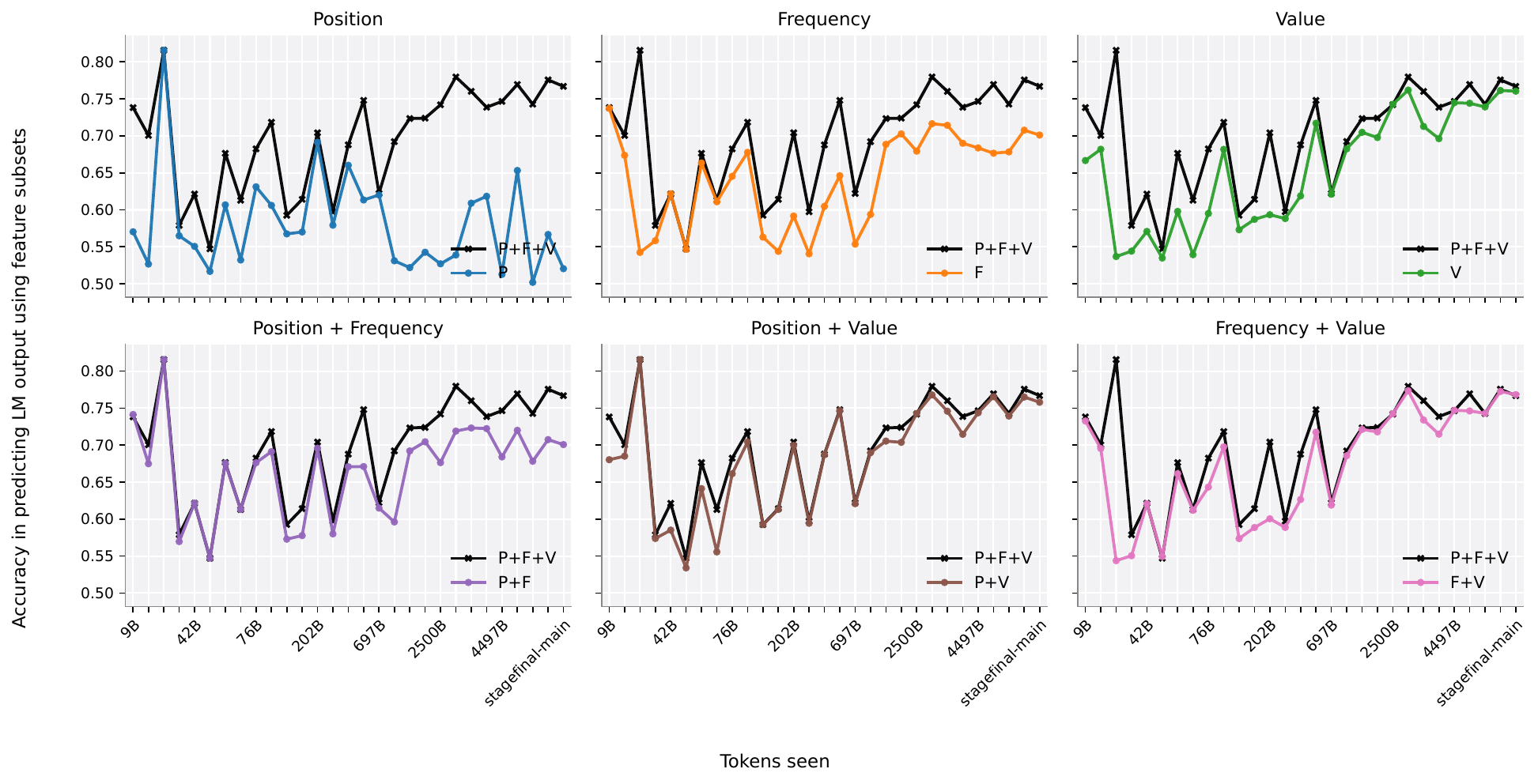}
  \caption{Prediction accuracy of feature subsets for OLMo-2 13B training checkpoints.}
  \label{fig:subset-olmo-13b}
\end{figure}

\begin{figure}[htbp]
  \centering
  \includegraphics[width=\linewidth]{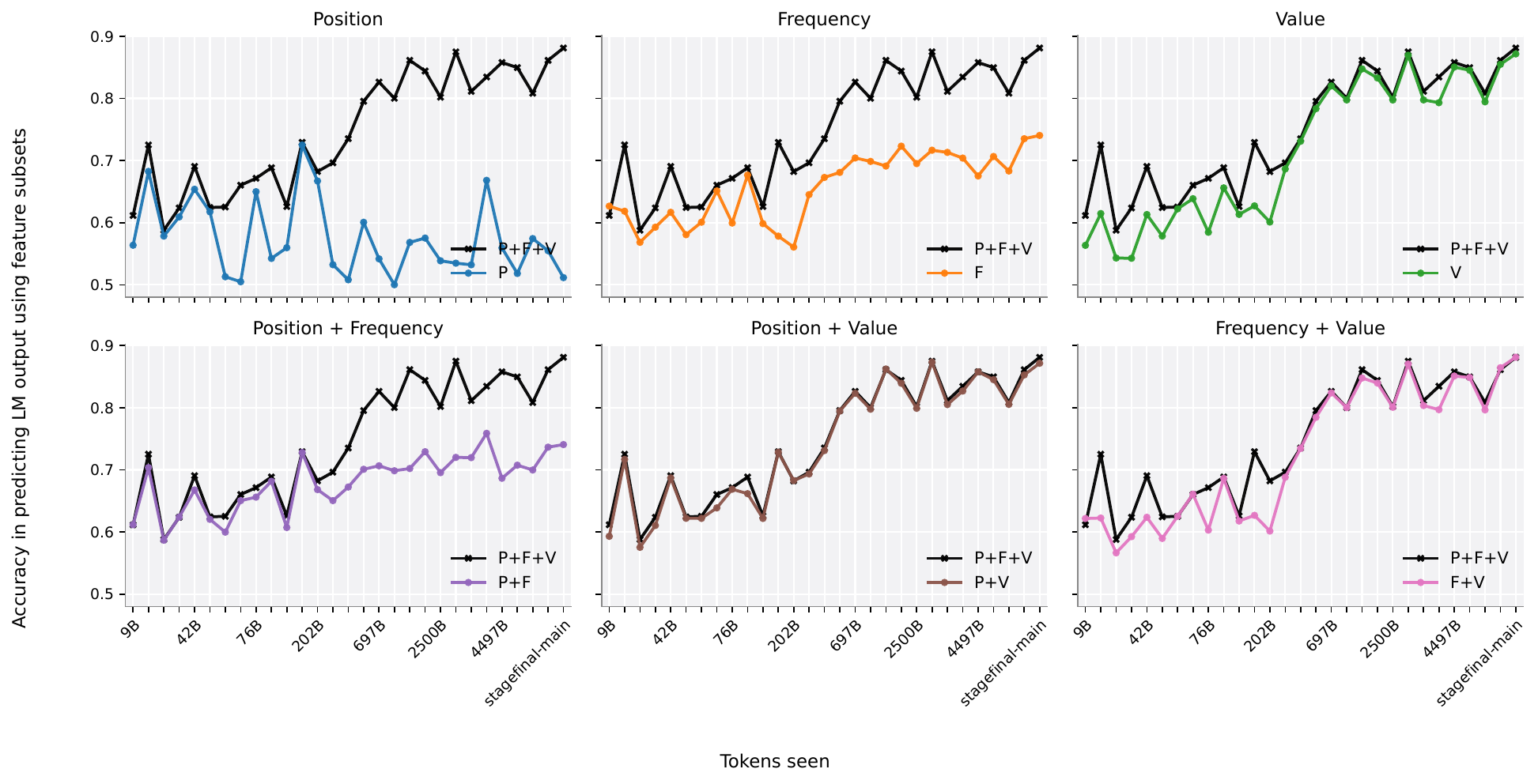}
  \caption{Prediction accuracy of feature subsets for OLMo-2 32B training checkpoints.}
  \label{fig:subset-olmo-32b}
\end{figure}

\FloatBarrier
\clearpage

\subsection{Evaluation on Holdout Data}\label{app:holdout}
\begin{figure}[htbp]
  \centering
  \includegraphics[width=0.7\linewidth]{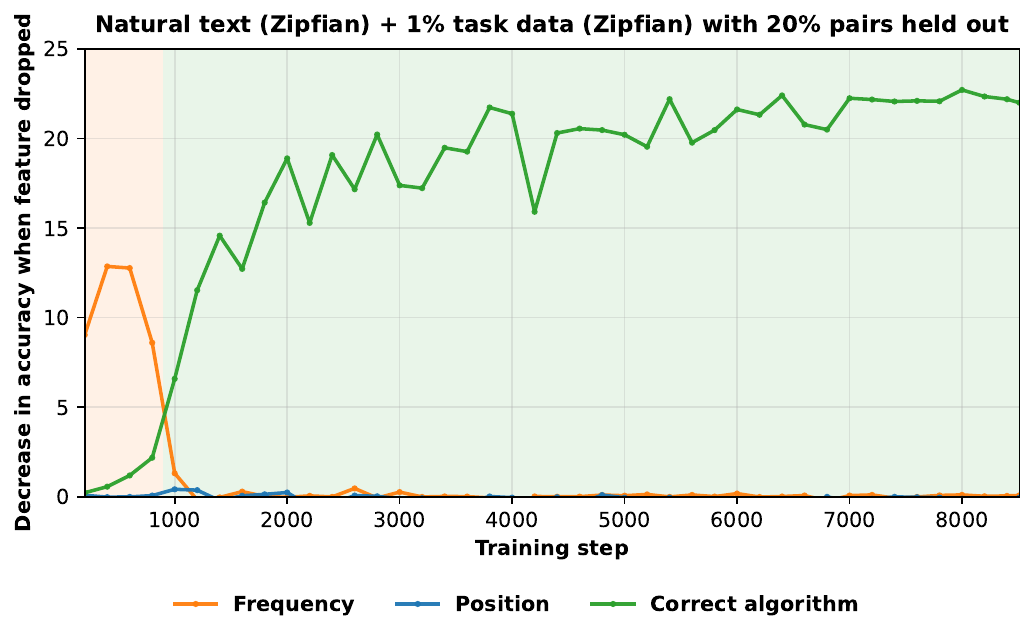}
  \caption{Holdout evaluation for the 163M small LM trained with 1\% Zipfian synthetic data, with 20\% of country pairs held out.}
  \label{fig:heldout-ablation}
\end{figure}

The controlled pretraining experiments in \Cref{sec:training} train and evaluate on all country entities.
To check that the correct algorithm phase is not an artifact of memorization, we additionally train a 163M small LM with 1\% Zipfian synthetic data, but restrict synthetic comparison statements to a random 80\% of country pairs. 
The remaining 20\% country pairs (4,644 out of 23,220 pairs) never appear in the synthetic data during training. 
We then evaluate on the holdout pairs only, so the model has never seen that specific comparison.
\Cref{fig:heldout-ablation} shows that the value feature rises in late training to a level comparable to that of in-distribution pairs. 

\section{GPT-2 Pretraining Experiments}\label{app:training}

The GPT-2-scale small LMs are trained on 4 NVIDIA H100 GPUs with bfloat16 mixed precision,  which takes roughly two hours for 7,560 training steps.
Each model is trained for two epochs over a mixture of about 1B tokens.
We save and evaluate a checkpoint every 200 steps.

Each model is trained with three random seeds (3407, 42, 123), which set the initialization and the order of the data.
We use AdamW ($\beta_1=0.9$, $\beta_2=0.95$, weight decay 0.1) with a learning rate that warms up linearly over 800 steps. Gradients are clipped to a norm of 1.0. 
The effective batch size is 256 sequences of 1024 tokens (16 sequences per GPU $\times$ 4 GPUs $\times$ 4 gradient-accumulation steps).
Each training sequence is of 1024 tokens.

\subsection{Training Data}\label{app:training-data}
Here, we describe the data compositions for training the small LMs in each controlled setup.

\paragraph{Varying the amount of synthetic data.}
For experiments in \Cref{sec:ratio}, we hold the data distributions of natural and synthetic text fixed to Zipfian (matching the \texttt{OLMo-mix-1124} distribution) and vary only the ratio of natural to synthetic text. To reach a target amount of synthetic data, we tile the synthetic comparison statements until they sum to the desired token count. We fill the rest of the token budget with natural text data sampled from \texttt{OLMo-mix-1124}. The exact token splits used in each run are listed in \Cref{tab:ratio-sweep}.

\begin{table}[htbp]
\small
\centering
\caption{Synthetic and natural text token splits for experiments in Section~\ref{sec:ratio}.}
\begin{tabular}{lrrr}
\toprule
\textbf{Run} & \textbf{Synth tokens / epoch} & \textbf{Natural tokens / epoch} \\
\midrule
0\%        & 0          & 1{,}000{,}000{,}288 \\
0.01\%     & 100{,}352  & 999{,}899{,}936     \\
0.1\%      & 1{,}000{,}448 & 998{,}999{,}840  \\
1\%        & 10{,}000{,}384 & 989{,}999{,}104 \\
\bottomrule
\end{tabular}
\label{tab:ratio-sweep}
\end{table}

\begin{figure}[htbp]
  \centering
  \includegraphics[width=\linewidth]{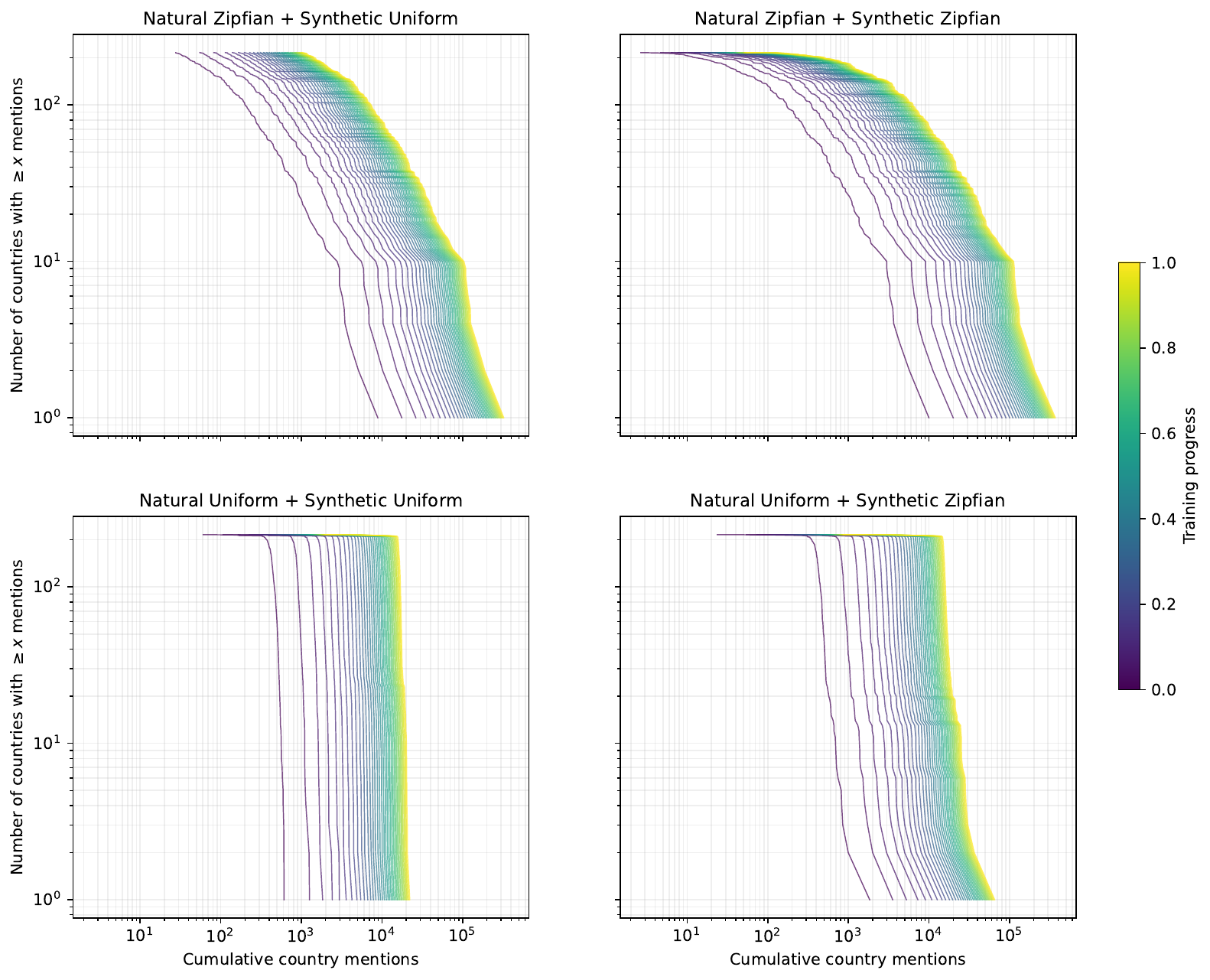}
  \caption{Visualizes the frequency distributions of country entities in each controlled pretraining condition. 
  For each panel, the x-axis is the number of times a country has been mentioned so far in training, and the y-axis is how many countries have been mentioned at least that many times (both axes on a log scale). Each line is one checkpoint, from early (dark purple) to late (yellow) training; lines move to the right as mentions accumulate. A line that falls gradually means countries differ widely in how often they are mentioned (a few are very frequent, many are rare). A line that drops almost vertically means nearly all countries have about the same number of mentions. 
  With Zipfian natural text (top row), the lines keep a gradual slope throughout training, and by the end the most frequent countries have a few hundred thousand mentions while many have far fewer. 
  With uniform natural text (bottom row), the lines are nearly vertical, and by the end almost all countries have about $10^4$ mentions. The synthetic data (left vs.\ right column) has a much smaller effect.}
  \label{fig:training-data-freq}
\end{figure}

\paragraph{Varying the distribution.}
For experiments in \Cref{sec:distribution}, we hold the ratio of natural to synthetic text fixed to 0.1\% and vary the entity frequency distributions of either natural text or synthetic data, one at a time. 
The distribution can be either Zipfian or uniform, which yields four conditions.
\Cref{tab:distribution-stats} reports summary statistics of the resulting combined distributions, and \Cref{fig:training-data-freq} visualizes the distribution in each condition.

\begin{table}[htbp]
  \centering
  \caption{Data distribution statistics across the four training conditions in \Cref{sec:distribution}. We compute the Gini ratio (0 = uniform, 1 = maximally unequal) and range ratios. Max/min is the count of the most mentioned country divided by the count of the least mentioned country, while P$_{95}$/P$_5$ uses the 95th and 5th percentile counts.}
  \label{tab:distribution-stats}
  \begin{tabular}{lccc}
  \toprule
  Condition & Gini coefficient & Max/min ratio & P$_{95}$/P$_5$ ratio \\
  \midrule
  \texttt{OLMo-mix-1124} (reference) & 0.772 & 8{,}600 & 470 \\
  \midrule
  Natural Zipfian + Synthetic Zipfian & 0.722 & 6{,}800 & 300 \\
  Natural Zipfian + Synthetic Uniform & 0.671 & 320 & 69 \\
  Natural Uniform + Synthetic Zipfian & 0.071 & 65 & 1 \\
  Natural Uniform + Synthetic Uniform & 0.031 & 11 & 1 \\
  \bottomrule
  \end{tabular}
\end{table}

\newpage
\FloatBarrier

\clearpage
\subsection{People Birth Year Comparison}\label{app:people_birthyear}
\begin{figure}[htbp]
  \centering
  \includegraphics[width=0.8\linewidth]{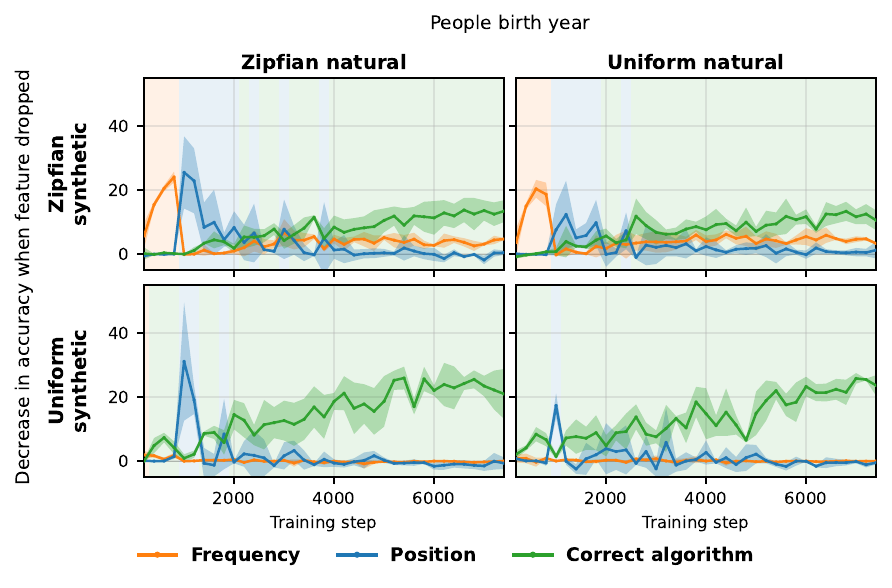}
  \caption{Learning trajectories for the people birth year task across four combinations of natural text and synthetic data distributions (Zipfian or uniform). Solid lines and shaded regions show mean and standard deviation across 3 seeds.}
  \label{fig:people_birth_year}
\end{figure}

Figure~\ref{fig:people_birth_year} shows the learning trajectories of the people birth year task under each of the four data mix conditions.
We observe that when the synthetic data is Zipfian, there is a strong early frequency phase until around step 1000, but this frequency phase becomes weaker when the synthetic data is uniform.
One possible explanation is that people are mentioned less often than countries in natural pretraining text (i.e., in the Zipfian natural text we use, the most and least frequently mentioned people have 9,598 and 0 mentions respectively, while the corresponding counts for countries are 165,998 and 26). As a result, the long tail in the synthetic data could affect learning more strongly.

\newpage
\FloatBarrier
\section{Evaluation on OLMo Models}\label{app:olmo}

\begin{table}[htbp]
\centering
\caption{OLMo-2 checkpoints used in our evaluation, with comparison accuracy at each checkpoint. We compare the logprobs of the first token at which the entity names diverge (e.g., when comparing \textit{United States} and \textit{United Kingdom}, we take the logprobs of \textit{States} and \textit{Kingdom}).}
\label{tab:olmo_checkpoints}
\small
\setlength{\tabcolsep}{3pt}
\resizebox{\textwidth}{!}{%
\begin{tabular}{rrr@{\hspace{0.8em}}rrr@{\hspace{0.8em}}rrr@{\hspace{0.8em}}rrr}
\toprule
\multicolumn{3}{c}{\textbf{OLMo-2 1B}} & \multicolumn{3}{c}{\textbf{OLMo-2 7B}} & \multicolumn{3}{c}{\textbf{OLMo-2 13B}} & \multicolumn{3}{c}{\textbf{OLMo-2 32B}} \\
\cmidrule(lr){1-3}\cmidrule(lr){4-6}\cmidrule(lr){7-9}\cmidrule(lr){10-12}
Step & Tokens & Acc. & Step & Tokens & Acc. & Step & Tokens & Acc. & Step & Tokens & Acc. \\
\midrule
       1,000 &        3B & 0.598 &        1,000 &        5B & 0.651 &        1,000 &        9B & 0.667 &        1,000 &        9B & 0.564 \\
       2,000 &        5B & 0.667 &        2,000 &        9B & 0.668 &        2,000 &       17B & 0.682 &        2,000 &       17B & 0.615 \\
       3,000 &        7B & 0.640 &        3,000 &       13B & 0.657 &        3,000 &       26B & 0.537 &        3,000 &       26B & 0.543 \\
       4,000 &        9B & 0.623 &        4,000 &       17B & 0.680 &        4,000 &       34B & 0.544 &        4,000 &       34B & 0.543 \\
       5,000 &       11B & 0.637 &        5,000 &       21B & 0.534 &        5,000 &       42B & 0.571 &        5,000 &       42B & 0.613 \\
       6,000 &       13B & 0.592 &        6,000 &       26B & 0.615 &        6,000 &       51B & 0.534 &        6,000 &       51B & 0.579 \\
       7,000 &       15B & 0.607 &        7,000 &       30B & 0.512 &        7,000 &       59B & 0.598 &        7,000 &       59B & 0.622 \\
       8,000 &       17B & 0.629 &        8,000 &       34B & 0.584 &        8,000 &       68B & 0.539 &        8,000 &       68B & 0.639 \\
       9,000 &       19B & 0.581 &        9,000 &       38B & 0.651 &        9,000 &       76B & 0.595 &        9,000 &       76B & 0.585 \\
      10,000 &       21B & 0.565 &       10,000 &       42B & 0.544 &       10,000 &       84B & 0.682 &       10,000 &       84B & 0.656 \\
      23,000 &       49B & 0.566 &       12,000 &       51B & 0.508 &       12,000 &      101B & 0.573 &       12,000 &      101B & 0.613 \\
      35,000 &       74B & 0.550 &       18,000 &       76B & 0.521 &       18,000 &      151B & 0.587 &       18,000 &      151B & 0.627 \\
      50,000 &      105B & 0.523 &       24,000 &      101B & 0.459 &       24,000 &      202B & 0.593 &       24,000 &      202B & 0.601 \\
      70,000 &      147B & 0.481 &       36,000 &      151B & 0.606 &       36,000 &      302B & 0.588 &       36,000 &      302B & 0.686 \\
     100,000 &      210B & 0.503 &       47,000 &      198B & 0.558 &       48,000 &      403B & 0.619 &       48,000 &      403B & 0.731 \\
     140,000 &      294B & 0.474 &       71,000 &      298B & 0.529 &       60,000 &      504B & 0.717 &       60,000 &      504B & 0.783 \\
     190,000 &      399B & 0.505 &       95,000 &      399B & 0.642 &       83,000 &      697B & 0.621 &       83,000 &      697B & 0.820 \\
     240,000 &      504B & 0.445 &      119,000 &      500B & 0.637 &      119,000 &      999B & 0.682 &      119,000 &      999B & 0.798 \\
     330,000 &      693B & 0.443 &      167,000 &      701B & 0.614 &      179,000 &    1,502B & 0.705 &      179,000 &    1,502B & 0.847 \\
     480,000 &    1,007B & 0.459 &      238,000 &      999B & 0.623 &      238,000 &    1,997B & 0.698 &      238,000 &    1,997B & 0.833 \\
     720,000 &    1,510B & 0.460 &      357,000 &    1,498B & 0.596 &      298,000 &    2,500B & 0.743 &      298,000 &    2,500B & 0.797 \\
     950,000 &    1,993B & 0.515 &      477,000 &    2,001B & 0.584 &      358,000 &    3,004B & 0.762 &      358,000 &    3,004B & 0.869 \\
   1,190,000 &    2,496B & 0.549 &      596,000 &    2,500B & 0.623 &      417,000 &    3,499B & 0.713 &      417,000 &    3,499B & 0.798 \\
   1,430,000 &    2,999B & 0.575 &      715,000 &    2,999B & 0.566 &      476,848 &    4,001B & 0.696 &      477,000 &    4,002B & 0.793 \\
   1,670,000 &    3,503B & 0.555 &      834,000 &    3,499B & 0.620 &      536,000 &    4,497B & 0.745 &      536,000 &    4,497B & 0.850 \\
   1,907,359 &    4,001B & 0.609 &      928,646 &    3,896B & 0.647 &      596,057 &    5,001B & 0.744 &      596,000 &    5,000B & 0.845 \\
\bottomrule
\end{tabular}%
}
\end{table}

\FloatBarrier
\clearpage

\subsection{Training OLMo-2 1B from Initialization}\label{app:olmo-early1000}
The earliest publicly released OLMo-2 checkpoint is already several billion tokens into training (e.g., the first OLMo-2 1B checkpoint, at step 1000, has already seen approximately 3B tokens), so the frequency phase documented in \Cref{sec:olmo} may be partially cut off. 
A stronger frequency phase could occur even earlier. To check this directly, we rerun the OLMo-2 1B stage-1 training recipe from random initialization until 1000 steps. 

\begin{table}[htbp]
\centering
\small
\caption{Dominant phase and drop-one accuracy (in parentheses, \%) for OLMo-2 1B trained from initialization, evaluated every 200 steps through the first 1000 steps of stage-1 pretraining, on all nine comparison tasks (larger magnitude variant).}
\label{tab:olmo-early1000}
\resizebox{\textwidth}{!}{%
\begin{tabular}{lrrrrr}
\toprule
\textbf{Task} & \textbf{Step 200} & \textbf{Step 400} & \textbf{Step 600} & \textbf{Step 800} & \textbf{Step 1000} \\
\midrule
Country area & Frequency (5.2) & Frequency (8.3) & Frequency (7.1) & Frequency (6.4) & Frequency (6.0) \\
Person birth year & Frequency (5.6) & Frequency (9.2) & Frequency (9.2) & Frequency (9.2) & Frequency (5.4) \\
Movie box office & Value (1.6) & Value (3.4) & Value (2.2) & Value (5.4) & Value (4.2) \\
Movie release year & Frequency (0.6) & Frequency (2.0) & Frequency (1.4) & Frequency (1.7) & Frequency (0.8) \\
State area & Frequency (21.3) & Frequency (11.9) & Frequency (12.4) & Frequency (10.9) & Frequency (8.3) \\
State population & Value (1.9) & Value (0.2) & Value (1.8) & Value (2.4) & Value (1.3) \\
University endowment & Frequency (7.7) & Frequency (8.3) & Frequency (6.3) & Frequency (5.2) & Frequency (5.7) \\
University founding year & Frequency (7.9) & Frequency (3.6) & Frequency (2.8) & Frequency (9.2) & Frequency (10.1) \\
University student population & Frequency (5.6) & Frequency (6.4) & Frequency (5.6) & Frequency (6.2) & Frequency (6.8) \\
\bottomrule
\end{tabular}%
}
\end{table}

\paragraph{The frequency heuristic is visible before the first public OLMo-2 checkpoint.} \Cref{tab:olmo-early1000} shows that seven of the nine tasks always show the frequency phase within the first 200 training steps, and remain so throughout this window.
Two tasks (movie box office and state population) instead show a value phase. The 1B model has low accuracy on these two tasks even at the end of training, so this likely reflects a noisy value-feature drop-one accuracy rather than genuine task competence.

\subsection{Results on All Comparison Tasks}\label{app:olmo-full}
\begin{figure}[htbp]
  \centering
  \includegraphics[width=0.9\linewidth]{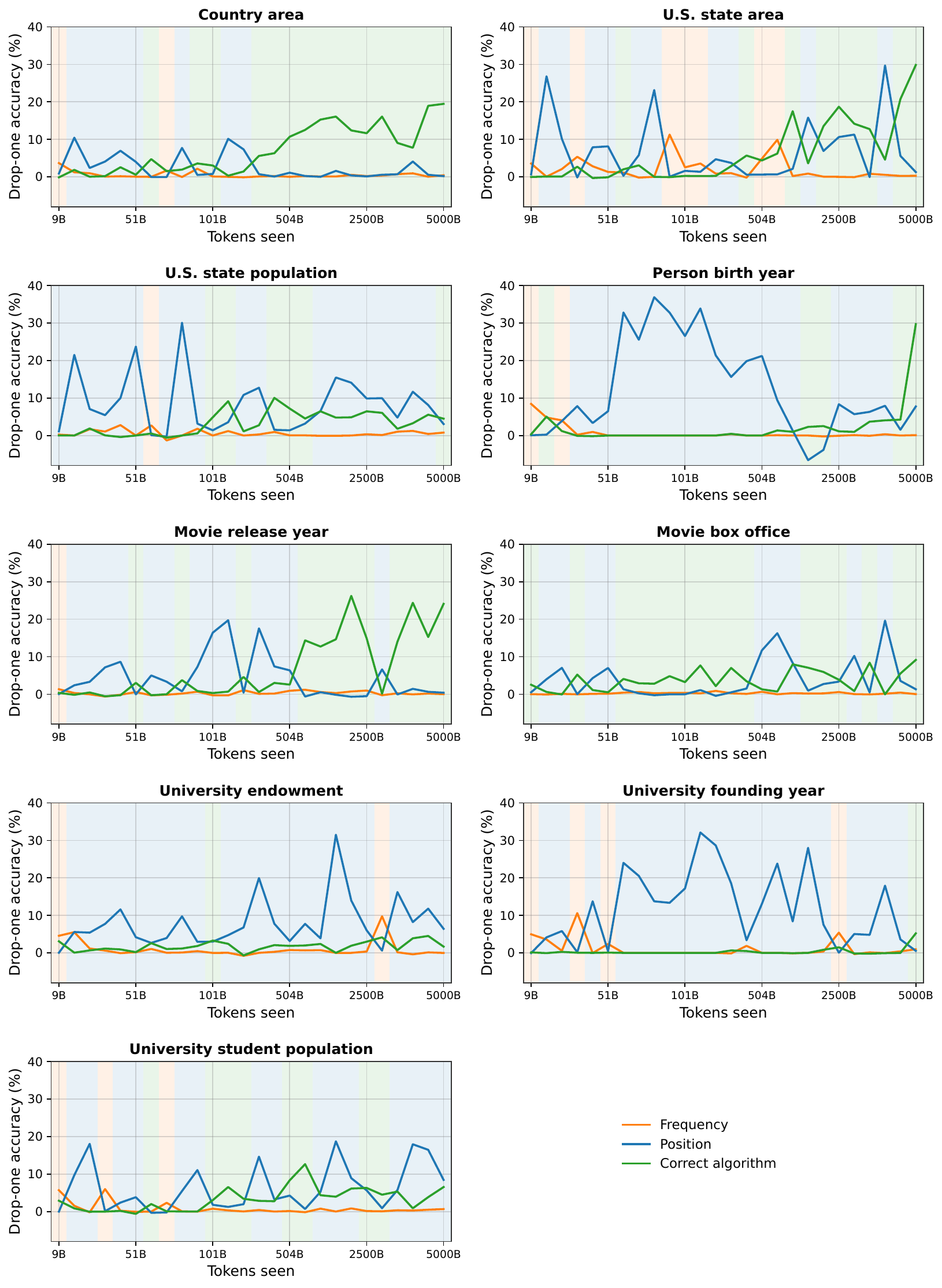}
  \caption{Learning trajectories for each of the nine comparison tasks for OLMo-2 32B, where the correct answer is the \textit{larger} entity.}
  \label{fig:olmo32b_all_larger}
\end{figure}

\begin{figure}[htbp]
  \centering
  \includegraphics[width=0.9\linewidth]{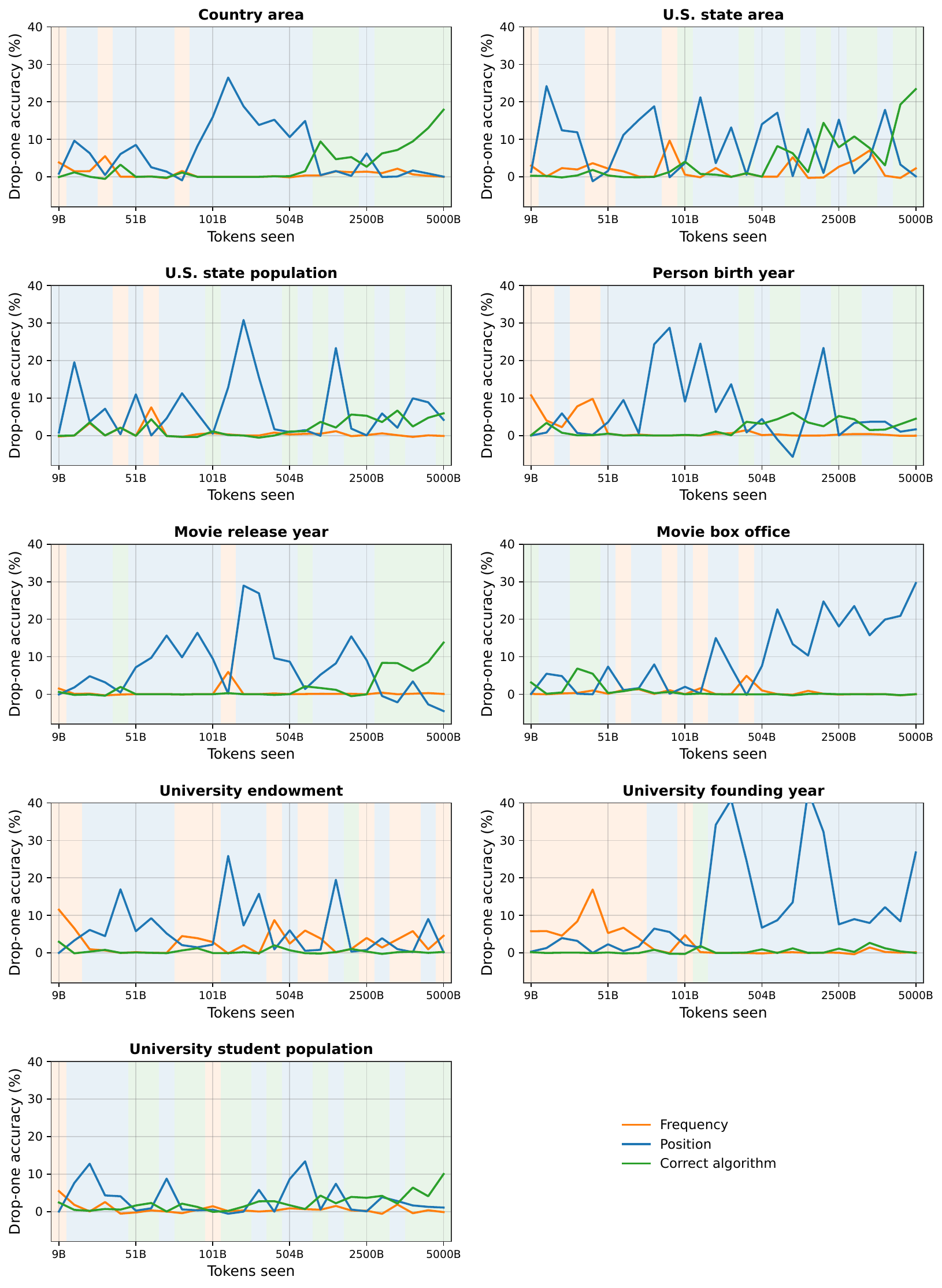}
  \caption{Learning trajectories for each of the nine comparison tasks for OLMo-2 32B, where the correct answer is the \textit{smaller} entity.}
  \label{fig:olmo32b_all_smaller}
\end{figure}

We analyze the learning trajectories of OLMo-2 32B on each of the nine comparison tasks. Figure~\ref{fig:olmo32b_all_larger} shows the features' drop-one accuracies when the correct answer is the larger entity, and Figure~\ref{fig:olmo32b_all_smaller} shows the results when the correct answer is the smaller entity.

\FloatBarrier
\clearpage


\end{document}